\documentclass{article}

    \PassOptionsToPackage{round}{natbib}

\usepackage[final]{neurips_2022}

\usepackage[utf8]{inputenc} 
\usepackage[T1]{fontenc}    
\usepackage[colorlinks=true,citecolor=blue,linkcolor=orange]{hyperref}       
\usepackage{url}            
\usepackage{booktabs}       
\usepackage{amsfonts}       
\usepackage{nicefrac}       
\usepackage{microtype}      
\usepackage{xcolor}         
\usepackage{xspace}

\usepackage{amssymb,amsmath}
\usepackage{amsthm}
\usepackage{cleveref}
\usepackage{bm}
\usepackage{mathtools}
\usepackage{todonotes}
\usepackage{float}

\newtheorem{theorem}{Theorem}

\graphicspath{{img/}}

\newcommand{\E}{\mathbb{E}}

\newcommand{\1}{\bm{I}}
\renewcommand{\vec}[1]{\bm{#1}}
\DeclareMathOperator{\KL}{KL}
\DeclareMathOperator{\smallkl}{kl}

\DeclareMathOperator*{\argmax}{argmax}
\DeclareMathOperator{\dirichlet}{Dir}
\DeclareMathOperator{\categorical}{Cat}
\DeclarePairedDelimiter\ceil{\lceil}{\rceil}
\newcommand{\measure}{\mathcal{M}_1^{+}}
\renewcommand{\Pr}{\mathbb{P}}
\renewcommand{\Re}{\mathbb{R}}

\newcommand{\aka}{{\it a.k.a.}\xspace}
\newcommand{\ie}{{\it i.e.}\xspace}

\renewcommand{\paragraph}{\textbf}
\allowdisplaybreaks    
\title{On Margins and Generalisation for Voting Classifiers}

\author{%
  Felix Biggs \\
  Department of Computer Science\\
  University College London and Inria\\
  London \\
  \texttt{contact@felixbiggs.com} \\
  \And
  Valentina Zantedeschi \\
  ServiceNow Research,\\
  University College London and Inria\\
  London \\
  \texttt{vzantedeschi@gmail.com} \\
  \AND
  Benjamin Guedj \\
  Department of Computer Science\\
  University College London and Inria\\
  London \\
  \texttt{b.guedj@ucl.ac.uk} \\
}

\begin{document}

\maketitle


\begin{abstract}
We study the generalisation properties of majority voting on finite ensembles of classifiers, proving margin-based generalisation bounds 
via the PAC-Bayes theory.
These provide state-of-the-art guarantees on a number of classification tasks.
Our central results leverage the Dirichlet posteriors studied recently by \citet{zantedeschi2021} for training voting classifiers;
in contrast to that work our bounds apply to non-randomised votes via the use of margins.
Our contributions add perspective to the debate on the ``margins theory'' proposed by \citet{schapireBoostingMarginNew1998} for the generalisation of ensemble classifiers.
\end{abstract}

\section{Introduction}

Weighted ensemble methods are among the most widely-used and effective
algorithms known in machine learning. Variants of boosting
\citep{DBLP:journals/jcss/FreundS97,DBLP:conf/kdd/ChenG16} are state-of-the-art
in a wide variety of tasks \citep{DBLP:journals/inffus/Shwartz-ZivA22,nielsen2016tree} and
methods such as random forest \citep{DBLP:journals/ml/Breiman01} are among the
most commonly-used in machine learning competitions \citep[see, \emph{e.g.},][]{netflix,uriot2021spacecraft}, valued both
for their excellent results and interpretability. Even when these algorithms do not directly produce the best learners for a task, the best performance in
competitions is often obtained by an ensemble of ``strong learners''---the output of a collection of different algorithms trained on the data---contrasted to the weak learners usually considered in the ensemble learning literature.

Among the oldest ideas to explain the performance of ensemble classifiers,
and machine learning methods in general, is the concept of margins. 
First introduced to analyse the Perceptron algorithm~\citep{novikoff62convergence}, 
margins relate closely to the idea of confidence in predictions in ensemble learning, 
with a large margin implying that a considerable weighted fraction of voters chose the same answer.
This was first leveraged to obtain early margin-based generalisation bounds for ensembles by
\citet{schapireBoostingMarginNew1998}, in an attempt to understand the excellent
generalisation of boosting, a surprising result given classical
Vapnik-Chervonenkis theory. This ``margins theory'' was explored further in a
number of works
\citep{DBLP:conf/colt/Wang08,DBLP:journals/ai/GaoZ13a,DBLP:conf/nips/GronlundKL20}
and is among the leading explanations for the success of such methods and
boosting in particular.

The same thread of margin bounds for ensemble methods has also been taken up in
parallel in PAC-Bayes theory by \citet{Langford01boundsfor,pmlr-v151-biggs22a}.
PAC-Bayes provides a natural framework both for deriving margin bounds, and for
considering ensemble methods in general, particularly majority votes where the
largest-weighted ensemble prediction is taken. Within the framework, the
weightings are typically considered as the parameter of a categorical
distribution over individual voters. PAC-Bayes theorems \citep[see the comprehensive surveys of][]{guedj2019primer,alquier2021userfriendly} then
directly provide generalisation bounds for the performance of this
``randomised'' proxy for the majority vote, \aka Gibbs classifier. 
These can then be de-randomised by such margin-based techniques, or through a variety of oracle bounds
\citep{langford2003pac,DBLP:journals/jmlr/Shawe-TaylorH09,DBLP:conf/pkdd/LacasseLMT10,DBLP:conf/nips/MasegosaLIS20},
motivating new learning algorithms
~\citep{DBLP:conf/nips/LacasseLMGU06,roy2011pac,germain2015risk,laviolette2017risk,lorenzen2019pac,viallard_Cbound,DBLP:journals/corr/abs-2106-13624}.

Uniquely among PAC-Bayesian approaches, \citet{zantedeschi2021} instead consider Dirichlet
distributions over the voters. Any sample from this distribution already implies
a vector of voting weights, and it is on the performance and optimisation of these
``stochastic majority votes'' they primarily focus. As an aside, they provide an
oracle result which allows their bounds to be de-randomised, but this introduces
an irreducible factor such that the bound on the true fixed vote can never be
less than double that of the stochastic version. It also neglects to leverage the
generally high confidence of predictions obtained by their algorithm.

\paragraph{Our contribution.}
By combining tools from margin bounds and the use of Dirichlet majority votes, we provide a new margin bound for non-randomised majority votes. This is in contrast to \citet{zantedeschi2021} which primarily considers stochastic majority votes.
Our bound empirically compares very favourably to existing margin
bounds and in contrast to them are applicable to multi-class classification.
Remarkably, our empirical results are also sharper than existing PAC-Bayesian ones, 
even when the algorithm optimising those bounds is used.

Our primary tool is a new result relating the margin loss of these stochastic votes to the misclassification loss of the non-randomised ones in a surprisingly sharp way.
This tool can additionally be utilised alongside a further idea from \citet{zantedeschi2021} to obtain an alternative form of the bound which is more amenable to optimisation.
Through this work we provide further support to the margins theory for
ensembles, showing that near-sharp bounds based on margins alone can be obtained
on a variety of real-world tasks.

\textbf{Outline.} The rest of this section introduces the problem setup, notation and summarises main results. \Cref{section:background} provides background on PAC-Bayes, Dirichlet majority votes and margin bounds, relating them to our new results.
\Cref{section:main-results} states and summarises our new theoretical results, giving the most relevant proofs (all remaining proofs are deferred to appendices).
\Cref{section:experiments} empirically evaluates these new results before we conclude with an overall discussion in \Cref{section:conclusion}.

\subsection{Notation and setting}\label{section:notation}

Majority voting algorithms combine the predictions of a finite set of ``base'' classifiers, \(\mathcal{H}\), from
\(\mathcal{X}\) to \(\mathcal{Y} = [c] := \{1, \dots, c\}\).
The classifiers \(h_i \in \mathcal{H}\) take the
form \(h_i: \mathcal{X} \to \mathcal{Y}\) for \(i \in [d]\) so that \(|\mathcal{H}| = d\).
Majority votes consider as set of
weightings \(\vec{\theta}\) in \(\Delta^d\), the simplex, and return the highest-weighted overall prediction.
Using the indicator function \(\1_{\mathcal{A}}\) of a set \(\mathcal{A}\), this is expressed as
\[f_{\vec{\theta}}(x) = \argmax_{k \in \mathcal{Y}} \sum_{i \in [d]} \theta_i \1_{h_i(x) = k}.\]

We are primarily interested in learning a weighting \(\vec{\theta}\) with small misclassification risk (and
guarantees of this) based on a sample \(S \sim \mathcal{D}^m\), where
\(\mathcal{D} \in \measure(\mathcal{X} \times \mathcal{Y})\) is the data-generating
distribution and \(m \in \mathbb{N}_{+}\) the sample size. We let \(\measure(\mathcal{A})\) denote the set of probability
measures on a set \(\mathcal{A}\). For \(h \in \mathcal{H}\) the misclassification loss is
\(\ell_0(h, x, y) := \1_{h(x) \ne y}\), the misclassification out-of-sample risk is
\(L_0(h) := \E_{(x, y) \sim \mathcal{D}} \ell_0(h, x, y)\) and a hat denotes the
in-sample estimate of this quantity,
\(\hat{L}_0(h) := \E_{(x, y) \sim \operatorname{Uniform}(S)} \ell_0(h, x, y)\). In a
slight abuse of notation we will also often write the risk of the majority vote
\(L_0(\vec{\theta}) = L_0(f_{\vec{\theta}})\) and similarly for its empirical counterpart.

The margin of majority vote \(f_{\vec{\theta}}\) on example \((x, y)\) is derived from the minimal gap between
the total weight assigned to the true class $y$ and to any other predicted class:
\[M(\vec{\theta}, x, y) := \frac12 \sum_{i : h_i(x) = y} \theta_i - \frac12 \max_{k \ne y} \sum_{i : h_i(x) = k} \theta_i.\]
The corresponding margin loss is
\(\ell_{\gamma}(\vec{\theta}, x, y) := \1_{M(\vec{\theta}, x, y) \le \gamma}\) for margin \(\gamma \ge 0\), with
the corresponding in-sample and out-of-sample risks notated as
\(\hat{L}_{\gamma}(\vec{\theta})\) and \(L_{\gamma}(\vec{\theta})\) respectively.

\subsection{Overview of results}

Our main result is a margin bound of the following form: with high
probability \(\ge 1 - \delta\) over the sample, simultaneously for any
\(\vec{\theta} \in \Delta^d\) and \(K > 0\),
\begin{equation}\label{eq:simplified-bound}
L(\vec{\theta}) \le O\left(\hat{L}_{\gamma}(\vec{\theta}) \; + \; e^{-K \gamma^{2}} \; + \; \frac{\mathbb{D}_{\dirichlet}(K \vec{\theta}, \bm{1}) + \log\frac{m}{\delta}}{m}\right)
\end{equation}
where \(\mathbb{D}_{\dirichlet}(\vec{\alpha}, \vec{\beta})\) is the KL
divergence between Dirichlet random vectors with parameters \(\vec{\alpha}\) and
\(\vec{\beta}\), with \(\bm{1}\) a vector of ones implying a uniform Dirichlet prior
distribution on the simplex. The term \(e^{-K \gamma^{2}}\) is a de-randomisation penalty. The
parameter \(K\) is chosen freely in an arbitrary data-dependent way to balance the
requirements of the different terms: it must be large enough to decrease this
exponential term, while too-large a parameter increases the KL divergence from
the uniform prior. This result is surprisingly strong; in
particular there is no dependence on the dimensionality (\ie, number of voters
$d$) in the exponential term, an advantage discussed further in
\Cref{section:main-proof}.

In \Cref{eq:simplified-bound}, \(\hat{L}_{\gamma}(\vec{\theta})\) is the 0-1 valued
\(\gamma\)-margin loss which enables comparison with existing margin bounds for trained
weighted ensembles. 
We further consider a second scenario, where the generalization bound is also used 
to train the model itself.
We note that the \(\gamma\)-margin loss \(\hat{L}_{\gamma}(\vec{\theta})\) appearing in \Cref{eq:simplified-bound} has null gradients,
so the bound
cannot be directly optimised by gradient descent. To rectify this we also prove a
variation of the bound, replacing the above loss by its expectation under a
Dirichlet stochastic vote,
\(\E_{\vec{\xi} \sim \dirichlet(K \vec{\theta})} \hat{L}_{\gamma}(\vec{\xi})\), which is bounded in differentiable closed form to give an alternative, optimisation-friendly bound.

In our evaluations we focus on these two complementary scenarios,
obtaining state-of-the-art empirical results.
Across different scenarios and tasks our results outperform both existing margin bounds (including a sharpened version of the result from \citet{pmlr-v151-biggs22a} which may be of independent interest), and PAC-Bayes bounds, even when it is not used as the objective.
Further, in contrast to existing margin bounds
our results also hold for multi-class majority votes.

\section{Background}\label{section:background}

\subsection{PAC-Bayes bounds}

PAC-Bayes bounds are among the tightest known generalisation bounds, as for
example the only framework in which non-vacuous generalisation bounds for neural
networks have been obtained \citep[see \emph{e.g.}][]{dziugaite2017computing,NIPS2018_8063,DBLP:conf/iclr/ZhouVAAO19,NIPS2019_8911,DBLP:journals/corr/abs-2006-10929,JMLR:v22:20-879,DBLP:journals/entropy/BiggsG21,biggs2022shallow}. However, unlike many
other such bounds they usually apply to randomised Gibbs(-like) prediction
functions rather than deterministic ones. These are typically re-drawn for every
new test evaluation. Thus a high-probability bound is obtained on the
expectation of the risk w.r.t. the PAC-Bayes posterior \(Q\), with the complexity of
\(Q \in \measure(\mathcal{H})\) appearing in the bound in terms of a Kullback-Leibler (KL) divergence from
a pre-chosen PAC-Bayes prior \(P \in \measure(\mathcal{H})\) \citep[which is not required to be a true prior in the
Bayesian sense -- see the discussion in][]{guedj2019primer}. A particularly sharp \citep[as discussed in][]{DBLP:conf/nips/FoongBBT21} and
widely-used result is given in \Cref{theorem:seeger}, valid for any bounded loss
function \(\ell\) with values in \([0, 1]\).

\begin{theorem}[\citet{seeger2001improved,DBLP:journals/corr/cs-LG-0411099}]\label{theorem:seeger}
For any \(\mathcal{D} \in \measure(\mathcal{X} \times \mathcal{Y})\), \(m \in \mathbb{N}_{+}\), prior \(P \in \measure(\mathcal{H})\) and
\(\delta \in (0, 1)\), with probability \(\ge 1 - \delta\) over \(S \sim D^{m}\), simultaneously for all
\(Q \in \measure(\mathcal{H})\)
\begin{equation*}
\E_{h \sim Q}L(h) \le \smallkl^{-1}\left(\E_{h \sim Q} \hat{L}(h), \; \frac{1}{m} \left( \KL(Q, P) + \log\frac{2\sqrt{m}}{\delta} \right) \right)
\end{equation*}
where the generalised inverse
\(\smallkl^{-1}(u, c) := \sup \{v \in [0, 1] : \smallkl(u, v) \le c\}\) and
\(\smallkl(u, v) := u \log\frac{u}{v} + (1-u)\log\frac{1-u}{1-v}\) is a KL
divergence between Bernoulli random variables.
\end{theorem}

The above bound uses the inverse small-kl function which will be seen in our
later results and a number of pre-existing ones. To lend intuition we
note that \(\smallkl^{-1}(u, c) \in O(u + c)\), giving
\Cref{eq:simplified-bound} from \Cref{theorem:main-bound} when using a uniform
prior. The following upper bounds are also useful:
\(\smallkl^{-1}(u, c) \le u + \sqrt{c/2}\) giving ``slow-rates'' and
\(\smallkl^{-1}(u, c) \le u + \sqrt{2c u} + 2c\). From this we can see that when the loss \(\hat{L} \to 0\) then the overall rate improves to \(O(1/m)\), so
the small-kl formulation interpolates between the traditional fast and slow rate
regimes of learning theory.

\subsection{Margin bounds}\label{section:background-margin-bounds}

In the learning theory literature there exists a rich tradition of using the
concept of a margin, which quantifies the confidence of predictions, to explain
generalisation. This is particularly evident in the case of voting algorithms
such as boosting, where traditional Vapnik-Chervonenkis based techniques predict
classical overfitting which is not ultimately observed. The ``margins theory''
was developed by \citet{schapireBoostingMarginNew1998} to explain this discrepancy. By considering the weightings \(\vec{\theta}\) as the parameter of a categorical distribution, they proved a bound of the
form (holding with probability greater than \(1 - \delta\) over the sample, as for all bounds in this section)
\(L_0(\vec{\theta}) \le L_{\gamma}(\vec{\theta}) + \tilde{O}\left(\frac{1}{\gamma \sqrt{m}}\right)\).
Although there was initially some debate about the validity of the theory
\citep{breiman1999}, eventually \citet[Theorem 4]{DBLP:journals/ai/GaoZ13a}
provided the following improved bound which further supported that a
large-margin voting classifier could generalise: simultaneously for any
\(\gamma > \sqrt{2/d}\) and \(\vec{\theta} \in \Delta^{d}\),
\begin{equation}\label{eq:kth-margin}
L_0(\vec{\theta}) \le \smallkl^{-1} \left(\hat{L}_{\gamma}(\vec{\theta}), \; \frac{1}{m} \left(\frac{2\log(2d)}{\gamma^2} \log\frac{2m^2}{\log d} + \log\frac{dm}{\delta}\right)\right) + \frac{\log d}{m}.
\end{equation}

More recently, a similar bound (proved through a PAC-Bayesian method based on
\citet{seeger2001improved}) was proved in \citet[Theorem 8]{pmlr-v151-biggs22a}.
Here we give a bound provided as an intermediate step in their proof that is
strictly (and empirically considerably) sharper than their final result: for any
fixed margin \(\gamma > 0\), simultaneously for any \(\vec{\theta} \in \Delta^{d}\)
\begin{equation}\label{eq:bgplus}
L_0(\vec{\theta}) \le \smallkl^{-1}\left(\hat{L}_{\gamma}(\vec{\theta}) + \frac{1}{m}, \; \frac{1}{m} \left(\ceil*{2 \gamma^{-2} \log m} \log d + \log \frac{2\sqrt{m}}{\delta}\right)\right) + \frac{1}{m}.
\end{equation}
Since \(\gamma \in (0, \frac12)\) for non-vacuous results, a union bound
argument can be used to extend the above to fixed-precision \(\gamma\), and this
result has the advantage of being valid for small \(\gamma\) as are often observed
empirically.

\paragraph{Our contributions.}
Firstly we mention the smaller contribution of the improved form of the bound from \citet{pmlr-v151-biggs22a} given in \Cref{eq:bgplus}; a proof is given in \Cref{section:more-margins-stuff} alongside further refinements and evaluation.
However we show that in many cases even this improved version and \Cref{eq:kth-margin} give weak or vacuous results.
As a result of this weakness (and thus perhaps null result for the margins theory applied to voting classifiers) we present a new margin bound in \Cref{theorem:main-bound} based on Dirichlet distributions as a theoretical intermediate step.
This is also valid in the multi-class case, unlike the above results which are only for binary classification.
Empirically the bound is observed to give an enormous improvement in tightness than the existing margin bounds and in some cases is near-sharp.

\subsection{Dirichlet stochastic majority votes}\label{section:background-dirichlet-mv}

In most results from the PAC-Bayes framework, and in the proof of the existing results given in \Cref{section:background-margin-bounds}, the majority vote weightings \(\vec{\theta}\) are considered the parameters of a categorical distribution over voters.
\citet{zantedeschi2021} instead consider PAC-Bayesian bounds (specifically,
\Cref{theorem:seeger}) applied to a hypothesis class of majority votes of the
form \(f_{\vec{\xi}}\), where \(\vec{\xi} \sim \dirichlet(\vec{\alpha})\) is drawn from a Dirichlet
distribution with parameter \(\vec{\alpha}\). This distribution has mean \(\E \vec{\xi} = \vec{\alpha}/\sum_{i=1}^d \alpha_i\) with a larger sum
\(\sum_{i=1}^d \alpha_i\)
giving a more concentrated or peaked distribution (see \Cref{section:about-dirichlet} for more details).

Since \(\vec{\xi}\) is randomised, the bounds from \citet{zantedeschi2021} apply to
``stochastic majority votes'' rather than the more typical deterministic ones we
consider here. However, the use of such Dirichlet distributions over voters in
the PAC-Bayes bounds rather than the more usual categorical ones is a major step
forward as it allows the correlation between voters to be more carefully
considered. This is because with a categorical distribution, the expected Gibbs risk is simply an average of the losses of individual predictors, without taking into account how well the combination of their predictions performs. Conversely, the Dirichlet distribution gives a (stochastic) majority vote of predictors, so if the errors of base voters are de-correlated, the better performance that results from their combination can be accounted for in the bound.
We will utilise and de-randomise these stochastic majority votes as a stepping stone to
bounds for deterministic predictors \(f_{\vec{\theta}}\) directly.

As is common in the PAC-Bayes literature, \citet{zantedeschi2021} use their new
bound as an optimisation objective to obtain a new algorithm, here using
stochastic gradient descent. The bound with Dirichlet posterior obtained
directly from \Cref{theorem:seeger} includes the expected misclassification loss
with respect to the Dirichlet parameters,
\(\E_{\vec{\xi} \sim \dirichlet(\vec{\alpha})}\ell(f_{\vec{\xi}}, x, y)\), which has null
gradient for any sampled \(\vec{\xi}\). They therefore additionally
upper bound this term by the differentiable closed form \begin{equation}\label{eq:incomplete-beta-bound}
\E_{\vec{\xi} \sim \dirichlet(\vec{\alpha})}\ell(f_{\vec{\xi}}, x, y) \le I_{\frac12}\left( \sum_{i : h_i(x) = y} \alpha_{i}, \sum_{i : h_i(x) \ne y} \alpha_{i} \right),
\end{equation}
where \(I_{z}(a, b)\) is the regularised incomplete beta function, which has a
sigmoidal shape. The inequality is sharp in the binary classification case, and
is used in the training objective and final evaluation of their method.
As an aside, \citet{zantedeschi2021} also proved an oracle bound which allows their result to be de-randomised, but this introduces a irreducible factor of two.
This bound, which holds with probability at least \(1{-}\delta\) over the sample for any \(\vec{\theta} \in \Delta^d, K > 0\) is given by
\begin{equation*}
L_0(\vec{\theta}) \le 2 \smallkl^{-1} \left(\E_{\vec{\xi} \sim \dirichlet(K \vec{\theta})}\hat{L}_0(\vec{\xi}), \; \frac{\mathbb{D}_{\dirichlet}(K \vec{\theta}, \vec{\beta}) + \log\frac{2\sqrt{m}}{\delta}}{m} \right).
\end{equation*}

\paragraph{Our contributions.} Firstly, we provide a new margin bound for majority vote algorithms utilising Dirichlet posteriors as a theoretical stepping stone.
We show that this bound gives sharper bounds on the misclassification loss than the bound from \citet{zantedeschi2021}, doing better than the irreducible factor, even when applied to the output of their algorithm.
We show further that the bound is also tighter when applied to the outputs of other PAC-Bayes algorithms derived from ``categorical''-type posteriors.
Finally, we give an altered form of the bound involving the expectation of the
margin loss
\(\E_{\vec{\xi} \sim \dirichlet(\vec{\alpha})}\ell_\gamma(f_{\vec{\xi}}, x, y)\) and a result analogous to \Cref{eq:incomplete-beta-bound} for this case.
Through this we are able to obtain a new PAC-Bayes objective which is compared to existing PAC-Bayes optimisation methods.

\section{Main results}\label{section:main-results}

Our main results use the idea of Dirichlet stochastic majority votes from
\citet{zantedeschi2021} as an intermediate step to prove new margin bounds for deterministic majority votes.
In this section, first we give our main result in \Cref{theorem:main-bound} and discuss further. In \Cref{section:bound-objective} we give an alternative bound obtained by a very similar method which is more amenable to optimisation, and we provide proofs for these results in \Cref{section:main-proof}.

The central step in these proofs is in constructing a proxy Dirichlet
distribution \(\vec{\xi} \sim \dirichlet(K\vec{\theta})\) over voters, the loss of which
is bounded à la PAC-Bayes, and de-randomised using margins to obtain bounds
directly for \(f_{\vec{\theta}}\).
The primary complexity term appearing in our bounds is therefore
\(\mathbb{D}_{\dirichlet}(K\vec{\theta}, \vec{\beta})\), the KL divergence between
Dirichlet distributions with parameters \(K\vec{\theta}\) and \(\vec{\beta}\)
respectively. As with PAC-Bayes priors, \(\vec{\beta}\) can be chosen in arbitrary
sample-independent fashion, but we typically choose it as a vector of ones,
giving a uniform distribution on the simplex as prior as in \Cref{eq:simplified-bound}.
The bounds also involve a de-randomisation penalty of
\(O(e^{-K\gamma^2})\) where \(\gamma\) is the margin appearing in the loss; this term
upper bounds the difference between our randomised proxy \(\vec{\xi}\) and its
mean \(\vec{\theta}\) and gets smaller with \(K\) as the distribution concentrates
tightly around its mean. This parameter \(K\) can be optimised in any data-dependent way to obtain the tightest final bound.

\begin{theorem}\label{theorem:main-bound}
For any \(\mathcal{D} \in \measure(\mathcal{X} \times \mathcal{Y})\), \(m \in \mathbb{N}_{+}\), margin \(\gamma > 0\), \(\delta \in (0, 1)\), and prior
\(\vec{\beta} \in \Re_{+}^d\), with probability at least \(1 - \delta\)
over the sample \(S \sim \mathcal{D}^m\) simultaneously for every \(\vec{\theta} \in \Delta^d\) and \(K > 0\),
\begin{equation*}
L_0(\vec{\theta}) \le \smallkl^{-1} \left(\hat{L}_{\gamma}(\vec{\theta}) + e^{-(K + 1) \gamma^{2}}, \; \frac{\mathbb{D}_{\dirichlet}(K \vec{\theta}, \vec{\beta}) + \log\frac{2\sqrt{m}}{\delta}}{m} \right) + e^{-(K + 1) \gamma^{2}}.
\end{equation*}
\end{theorem}

\Cref{theorem:main-bound} differs from the existing margin bounds of
\Cref{eq:bgplus,eq:kth-margin}, and
\citet{schapireBoostingMarginNew1998} in a specific and significant way, with
\(\vec{\theta}\) appearing not only in the loss function \(\hat{L}_{\gamma}(\vec{\theta})\),
but \emph{also} in the KL complexity term. Empirically we find our bound to be an improvement
but it is possible to generate scenarios where the pre-existing
bounds are non-vacuous while ours is not, since the KL divergence is unbounded
for certain choices of \(\vec{\theta}\), for example when one of the components is
exactly zero. This difference arises because the existing bounds all use the idea
of a categorical distribution with parameter \(\vec{\theta}\) in their proofs (which has KL divergence from a uniform prior upper bounded by \(\log d\)), while
we use a Dirichlet. This gains us the surprisingly tight de-randomisation result (\Cref{theorem:dirichlet-margin}) used in all proofs.

\subsection{PAC-Bayes bound as objective}\label{section:bound-objective}

We note here that it is non-trivial to directly obtain a training objective for optimisation from \Cref{theorem:main-bound}, due to the non-differentiability of the margin loss \(\hat{L}_{\gamma}(\vec{\theta})\). Therefore, in order to compare results with a wide variety of methods that optimise PAC-Bayes bounds \citep[including those used by][as baselines]{zantedeschi2021}, we obtain a relaxed and differentiable formulation in \Cref{theorem:main-bound-stochastic} for direct optimisation.

\begin{theorem}\label{theorem:main-bound-stochastic}
  Under the conditions of \Cref{theorem:main-bound} the following bound also holds
    \begin{equation*}
    L_0(\vec{\theta}) \le \smallkl^{-1} \left(\E_{\vec{\xi} \sim \dirichlet(K \vec{\theta})}\hat{L}_{\gamma}(\vec{\xi}), \; \frac{\mathbb{D}_{\dirichlet}(K \vec{\theta}, \vec{\beta}) + \log\frac{2\sqrt{m}}{\delta}}{m} \right) + e^{-4 (K + 1) \gamma^{2}}.
    \end{equation*}
    Using the incomplete Beta function \(I_z(a,b)\) we also have the following result, which is sharp in the binary classification case,
    \begin{equation*}
    \E_{\vec{\xi} \sim \dirichlet(\vec{\alpha})} \ell_{\gamma}(\vec{\xi}, x, y) \le I_{\frac12+\gamma}\left(\sum_{i : h_i(x) = y}\alpha_i, \sum_{i : h_i(x) \neq y}\alpha_i\right).
    \end{equation*}
\end{theorem}

\Cref{theorem:main-bound-stochastic} has a stronger PAC-Bayesian flavour than \Cref{theorem:main-bound}, with an expected loss under some distribution appearing
(complicating the final optimisation of \(K\)), while \Cref{theorem:main-bound} takes a
form much closer to that of a classical margin bound.
The second part of the result is analogous to \Cref{eq:incomplete-beta-bound} used by \citet{zantedeschi2021}.
We combine both parts to calculate the overall bound in closed form and obtain gradients for optimisation.

\subsection{Proof of main results}\label{section:main-proof}

The proof of \Cref{theorem:main-bound,theorem:main-bound-stochastic} essentially follow from applying a
simple PAC-Bayesian bound in combination with the key
\Cref{theorem:dirichlet-margin} below. In some sense this is our most important and novel result. Our whole approach is largely motivated
by its surprising tightness; in particular there is no dependence
on the dimension, which is avoided by careful use of the aggregation property of
the Dirichlet distribution. This surprise arises because to obtain a tightly
concentrated Dirichlet distribution on \(\vec{\xi} \sim \dirichlet(\vec{\alpha})\), the
concentration parameter \(K = \sum_{i=1}^d \alpha_i\) must grow linearly with the
dimension. In fact, even a uniform distribution (which will be less peaked than
our final posterior) has \(\sum_{i=1}^d \alpha_i = d\), so the de-randomisation step is
effectively very cheap in higher dimensions.

\begin{theorem}\label{theorem:dirichlet-margin}
Let \(\vec{\theta} \in \Delta^d\) and \(K > 0\). Then for any \(\gamma > 0\) and \((x, y)\),
\begin{align*}
\ell_0(\vec{\theta}, x, y) \le \E_{\vec{\xi} \sim \dirichlet(K\vec{\theta})}\ell_{\gamma}(\vec{\xi}, x, y) &+ e^{-4 (K + 1) \gamma^2},\\
\E_{\vec{\xi} \sim \dirichlet(K\vec{\theta})} \ell_\gamma(\vec{\xi}, x, y) \le \ell_{2\gamma}(\vec{\theta}, x, y) &+ e^{-4 (K + 1) \gamma^2}.
\end{align*}
\end{theorem}

For our proofs we first recall the
aggregation property of the Dirichlet distribution: if
\((\xi_1, \dots, \xi_d) \sim \dirichlet((\alpha_1, \dots, \alpha_d))\), then
\((\xi_1, \dots, \xi_{d-1} + \xi_d) \sim \dirichlet((\alpha_1, \dots, \alpha_{d-1} + \alpha_d))\).
We further note the following crucial concentration-of-measure result. The
aforementioned lack of dimensionality in \Cref{theorem:dirichlet-margin} is
possible because \Cref{marchal_arbel} depends only on \(\sum_{i=1}^d \alpha_i\), and
this value is unchanged by aggregation, which avoids the dimension dependence
that would otherwise be introduced by the requirement \(\|\vec{u}\|_2 = 1\)
below.

\begin{theorem}[\citealp{marchal2017sub}]\label{marchal_arbel}
Let \(\vec{X} \sim \dirichlet(\vec{\alpha})\), \(t > 0\), and \(\vec{u} \in \Re^d\) with
\(\|\vec{u}\|_2 = 1\). Then
\[ \Pr_{\vec{X}}\left\{ \vec{u} \cdot (\vec{X} - \E \vec{X}) > t \right\} \le \exp\left(-2\left(\sum_{i=1}^d \alpha_i + 1\right) t^2\right).\]
\end{theorem}

\begin{proof}[Proof of \Cref{theorem:main-bound} and \Cref{theorem:main-bound-stochastic}.]
The proof of our main results is completed by applying the PAC-Bayes
bound \Cref{theorem:seeger} with the \(\gamma\)-margin loss to a Dirichlet prior
and posterior with parameters \(\vec{\beta}\) and \(K\vec{\theta}\) respectively.
Substituting the first part of \Cref{theorem:dirichlet-margin} gives the first part of
\Cref{theorem:main-bound-stochastic}, and additionally substituting the second part
and re-scaling \(\gamma \to \gamma/2\) gives \Cref{theorem:main-bound}.

For the second part of \Cref{theorem:main-bound-stochastic}, define \(w = \{i : h_i(x) = y\}\) for fixed \((x, y)\) so
\(W := \sum_{i \in w} \xi_i \sim \operatorname{Beta}\left(\sum_{i \in w} \alpha_i, \sum_{i \notin w} \alpha_i\right)\)
by the aggregation property of the Dirichlet distribution. Then
\[\E_{\vec{\xi}} \ell_{\gamma}(\vec{\xi}, x, y) \le \E_{\vec{\xi}} \left\{W \ge \frac12 - \gamma\right\} = 1 - I_{\frac12 - \gamma}\left(\sum_{j \in w} \alpha_{j}, \sum_{i \notin w} \alpha_{j}\right) = I_{\frac12 + \gamma}\left(\sum_{j \notin w} \alpha_{j}, \sum_{i \in w} \alpha_{j}\right)\]
using \(\1_{a_i - \max_{j\ne i} a_j \le 2\gamma} \le \1_{a_i - \sum_{j\ne i} a_j \le 2\gamma} = \1_{\sum_{j \ne i} a_j \ge \frac12 - \gamma}\) for \(\vec{a} \in \Delta^c\) (with equality for \(c = 2\) classes), and that \(I_z(a, b)\) is the CDF of a Beta distribution with parameters \((a, b)\).
\end{proof}

\begin{proof}[Proof of \Cref{theorem:dirichlet-margin}.]
Define \(\gamma_2 > \gamma_1\) such that \(\gamma := \gamma_2 - \gamma_1\), and \(\vec{\alpha} = K \vec{\theta}\).
From the trivial inequality
\(\1_{x \in A} - \1_{x \in B} \le \1_{x \in A} \1_{x \notin B}\) we derive
\begin{align*}
\Delta &:= \ell_{\gamma_1}(\vec{\theta}, x, y) - \E_{\vec{\xi} \sim \dirichlet(\vec{\alpha})}\ell_{\gamma_2}(\vec{\xi}, x, y)
\; = \; \E_{\vec{\xi} \sim \dirichlet(\vec{\alpha})}[\1_{M(\vec{\theta}, x, y) \le 0} - \1_{M(\vec{\xi}, x, y) \le \gamma}] \\
&\le  \E_{\vec{\xi} \sim \dirichlet(\vec{\alpha})}[\1_{M(\vec{\theta}, x, y) \le 0} \1_{M(\vec{\xi}, x, y) > \gamma}]
\; \le \;  \E_{\vec{\xi} \sim \dirichlet(\vec{\alpha})}[\1_{M(\vec{\xi}, x, y) - M(\vec{\theta}, x, y) > \gamma}] \\
&= \Pr_{\vec{\xi} \sim \dirichlet(\vec{\alpha})}\left\{
\sum_{i : h_i(x) = y} \xi_i - \max_{j' \ne y} \sum_{i : h_i(x) = k'} \xi_i - \sum_{i : h_i(x) = y} \theta_i + \max_{j \ne y} \sum_{i : h_i(x) = k} \theta_i > 2 \gamma \right\} \\
&\le \Pr_{\vec{\xi} \sim \dirichlet(\vec{\alpha})}\left\{
\sum_{i : h_i(x) = y} \xi_i - \sum_{i : h_i(x) = k} \xi_i - \sum_{i : h_i(x) = y} \theta_i + \sum_{i : h_i(x) = k} \theta_i > 2 \gamma \right\} \\
\end{align*}
where in the last inequality we set
\(k = \operatorname{argmax}_{k \ne y} \sum_{i : h_i(x) = k} \theta_i\), and use that
\(\max_j \sum_{i:h_i(x) = j} \theta_i - \max_j \sum_{i:h_i(x) = j} \xi_i \le \max_j \sum_{i:h_i(x) = j} \theta_i - \sum_{i:h_i(x) = k} \xi_i\)
for any \(k\). We rewrite the above in vector form (with inner product denoted \(\vec{u} \cdot \vec{v}\)) as
\begin{align*}
\Delta
&\le \Pr_{\vec{\xi} \sim \dirichlet(\vec{\alpha})}\Biggl\{
\underbrace{\frac{1}{\sqrt{2}}
\begin{bmatrix} 1 \\ -1 \\ 0 \end{bmatrix}}_{\vec{u}} \; \cdot \;
\Biggl(
\underbrace{\begin{bmatrix}
\sum_{i : h_i(x) = y} \xi_i \\
\sum_{i : h_i(x) = k} \xi_i \\
\sum_{i : h_i(x) \notin \{k,y\}} \xi_i \\
\end{bmatrix}}_{\vec{\tilde{\xi}}}
-
\underbrace{\begin{bmatrix}
\sum_{i : h_i(x) = y} \theta_i \\
\sum_{i : h_i(x) = k} \theta_i \\
\sum_{i : h_i(x) \notin \{k,y\}} \theta_i \\
\end{bmatrix}}_{\E\vec{\tilde{\xi}}}
\Biggr)
> \sqrt{2}\gamma \Biggr\} \\
&= \Pr_{\tilde{\vec{\xi}} \sim \dirichlet(\tilde{\vec{\alpha}})}\left\{\vec{u} \; \cdot \; (\tilde{\vec{\xi}} - \E \tilde{\vec{\xi}}) > \sqrt{2}\gamma \right\}
\end{align*}
where by the aggregation property of the Dirichlet distribution
\(\vec{\tilde{\xi}} \sim \dirichlet(\tilde{\vec{\alpha}})\) with
\[ \vec{\tilde{\alpha}} := \left[\sum_{i : h_i(x) = y} \alpha_i, \; \sum_{i : h_i(x) = k} \alpha_i, \; \sum_{i : h_i(x) \notin \{k,y\}} \alpha_i\right]^{T}. \]

Applying \Cref{marchal_arbel} we obtain
\(\Delta \le e^{-4 (\sum_i \tilde{\alpha}_i + 1) \gamma^2} = e^{-4 (\sum_{i=1}^d \alpha_i + 1) \gamma^2}\).
This gives the first inequality by setting \(\gamma_1 = 0, \gamma_2 = \gamma\). Setting
\(\gamma_1 = \gamma, \gamma_2 = 2\gamma\) and swapping \(\vec{\theta}\) and \(\vec{\xi}\) gives an almost
identical proof (with some signs reversed) of the second inequality.
\end{proof}

\section{Empirical evaluation}\label{section:experiments}

\begin{figure}[t]
\centering
\includegraphics[width=0.9\textwidth]{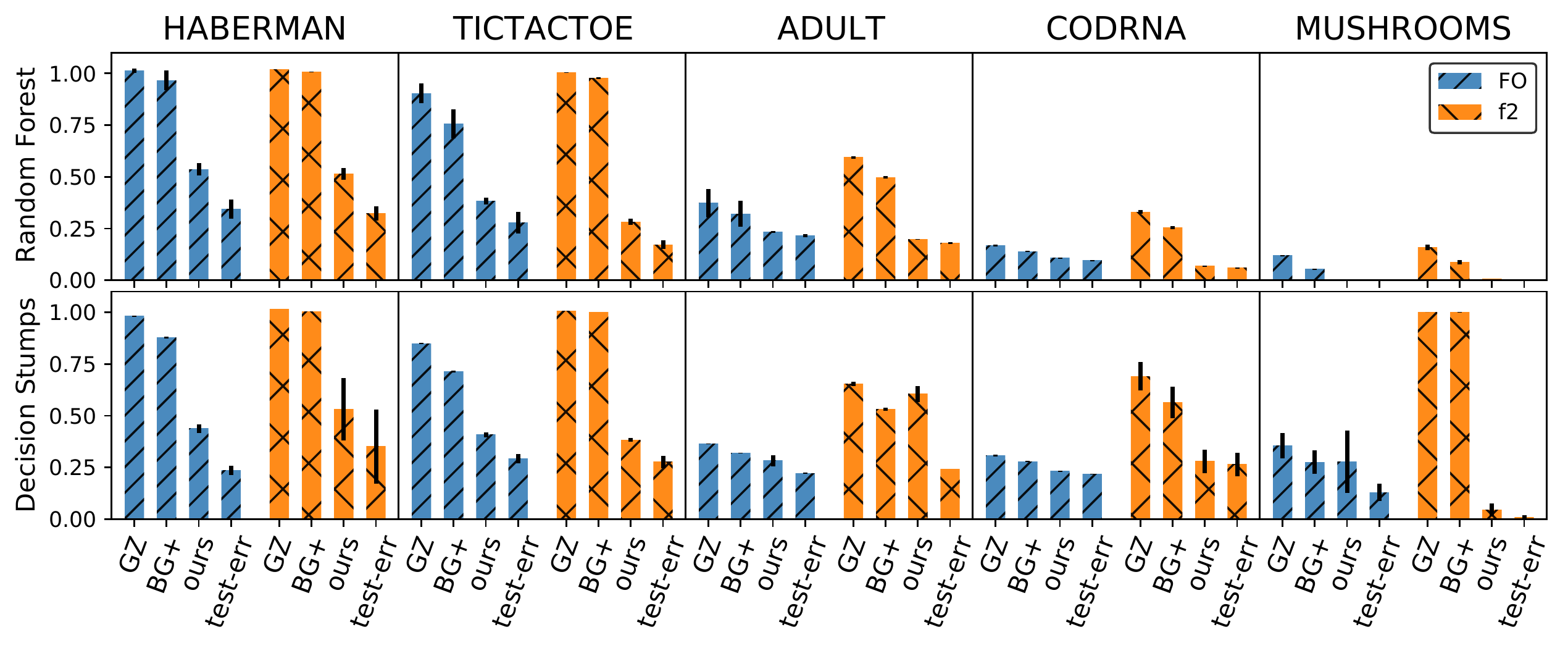}
\caption{\Cref{theorem:main-bound} (\textbf{ours}) compared with the margin bounds of \Cref{eq:bgplus} (\textbf{BG+}), \Cref{eq:kth-margin} (\textbf{GZ}), and the test error. Settings are \emph{rf} (first row) and \emph{stumps} (second row) on the given datasets, with \(\vec{\theta}\) output by optimising either \emph{FO} or \emph{f2} (first and second column groupings respectively).}
\label{fig:margins}
\end{figure}

\begin{figure}[t]
\centering
\includegraphics[width=0.9\textwidth]{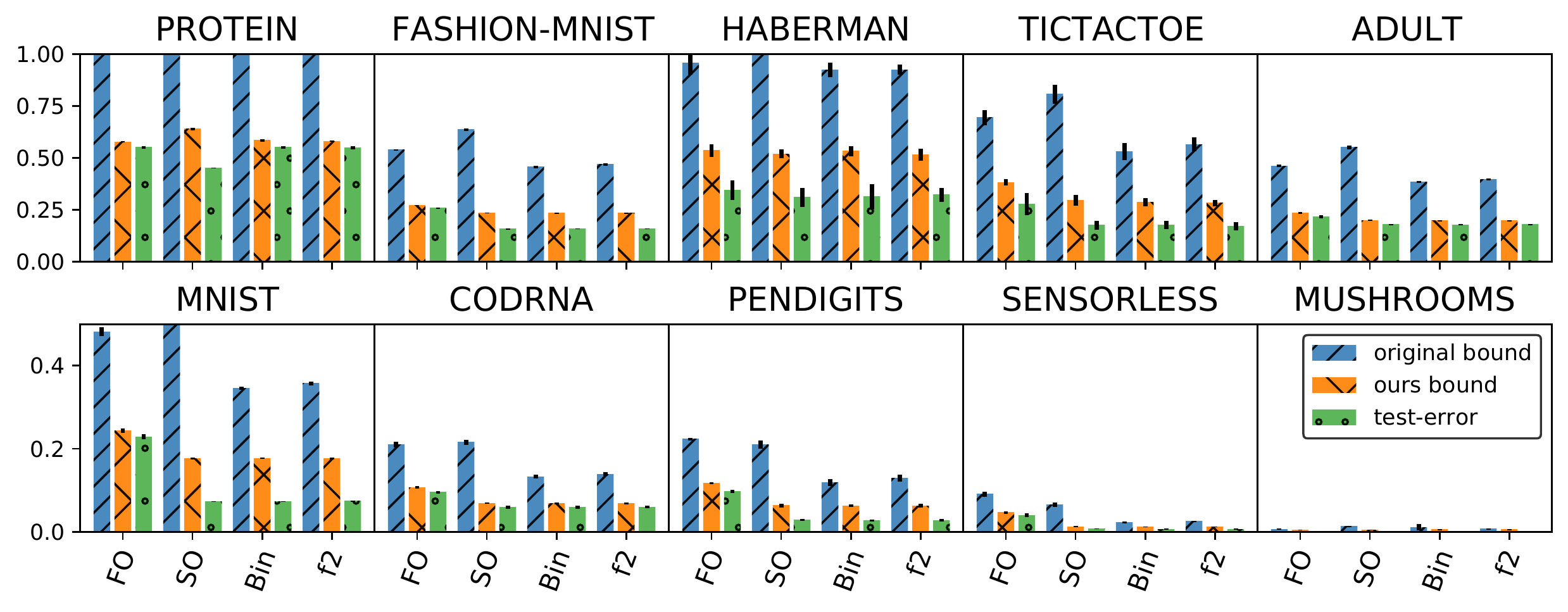}
\caption{\Cref{theorem:main-bound} (\emph{our bound}) compared with the bounds of  \emph{FO}, \emph{SO}, \emph{Bin} or \emph{f2} (\emph{original bound}), and test errors. For each column grouping, \(\vec{\theta}\) is the output from optimising the corresponding PAC-Bayes bound (as named underneath) for \emph{rf} on the given dataset. The blue column is the final value of the bound used as objective, the green is the test error, and the orange is the value of our bound when \(\vec{\theta}\) is plugged into it (so that our bound is not used as an objective here).}
\label{fig:pac-bayes}
\end{figure}

\begin{figure}[t]
\centering
\includegraphics[width=0.9\textwidth]{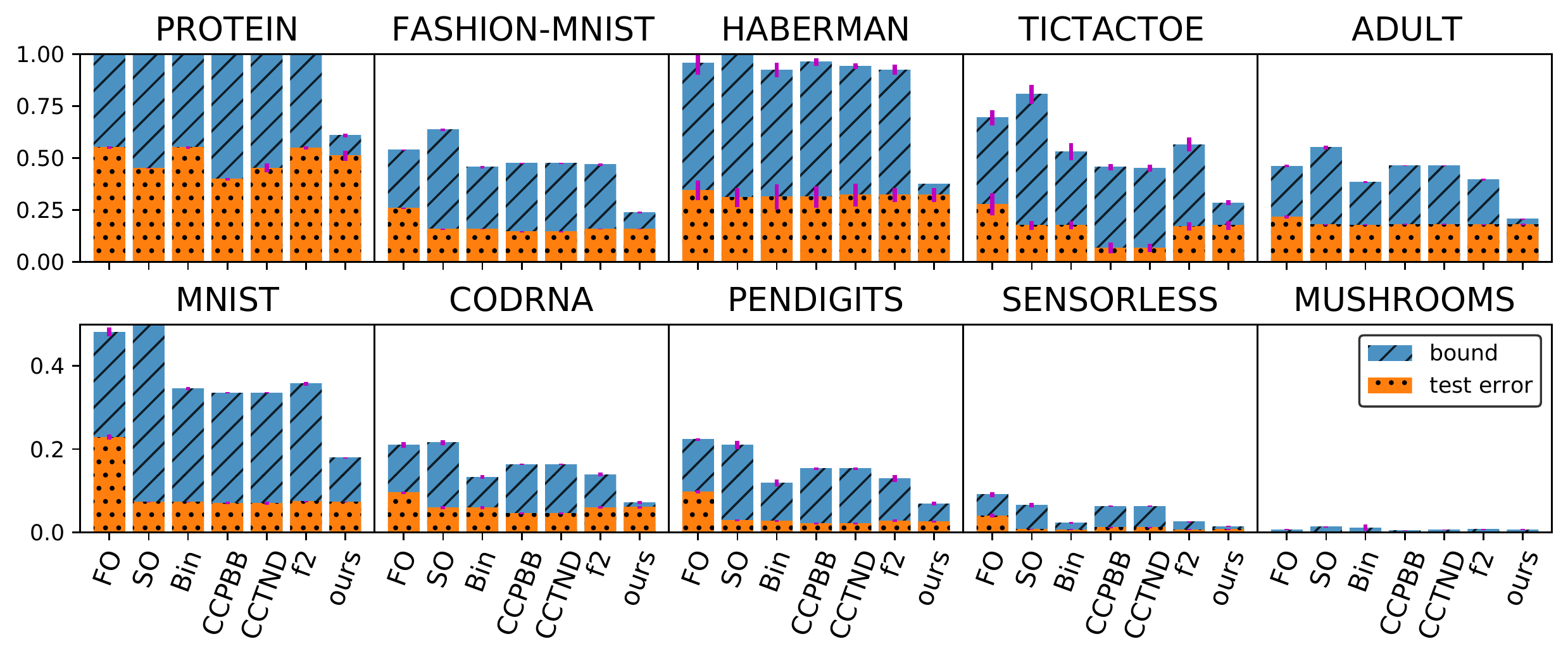}
\caption{\Cref{theorem:main-bound-stochastic} (\textbf{ours}) as optimisation objective compared to other PAC-Bayes results (\emph{FO}, \emph{SO}, \emph{Bin}, \emph{CCPBB} and \emph{CCTND}) as objectives in the \emph{rf} setting. For each objective the test error and bound associated with the objective is shown.}
\label{fig:optimisation}
\end{figure}

In this section we empirically validate our results against existing PAC-Bayesian and margin bounds on
several classification datasets from UCI~\citep{Dua2019}, LIBSVM\footnote{\url{https://www.csie.ntu.edu.tw/~cjlin/libsvm/}} and Zalando~\citep{xiao2017}.
Since our main result in \Cref{theorem:main-bound} is not associated with any particular algorithm, we
use \(\vec{\theta}\) outputted from
PAC-Bayes-derived algorithms to evaluate this result against other margin bounds (\Cref{fig:margins}) and PAC-Bayes bounds (\Cref{fig:pac-bayes}). 
We then compare optimisation of our secondary result \Cref{theorem:main-bound-stochastic} with optimising those PAC-Bayes bounds directly (\Cref{fig:optimisation}).
All generalisation bounds given are evaluated with a probability
\(1 {-} \delta {=} 0.95\). Further details not provided here including tabulated
results, description of datasets, training mechanisms and compute are provided
in \Cref{section:additional-empirical}.
The code for reproducing the results is available at \url{https://github.com/vzantedeschi/dirichlet-margin-bound}.

\paragraph{Strong and weak voters.} Similarly to \citet{zantedeschi2021}
we consider both using data-independent and data-dependent voters. This brings our experimental
setup in line with a common workflow for machine learning practicioners: the
training set is sub-divided into a set for training several different strong
algorithms, and a second set on which the weightings of these are optimised.
More specifically, the weak voter setting, used only for binary classification, uses
axis-aligned decision stumps (denoted \emph{stumps}), with thresholds evenly spread over the input space
($6$ per feature and per class). The stronger voters (denoted \emph{rf}) are learned from half of
the training data, while the other half is used for evaluating and optimising
the different generalisation bounds (note this reduces \(m\)). These take the
form of random forests~\citep{DBLP:journals/ml/Breiman01} of $M{=}10$ trees optimising
Gini impurity score on $\frac{n}{2}$ bagged samples and $\sqrt{d}$ drawn
features for each tree, with unbounded maximal depth.

\paragraph{Optimising \(\gamma\) and \(K\) in bounds.} In reporting margin bounds we
optimise over a grid of margin \(\gamma\) values in \((0, \frac12)\), and
additionally over \(K\) for \Cref{theorem:main-bound}. Since
\Cref{theorem:main-bound} and \Cref{eq:bgplus} as stated require a fixed margin,
we apply a union bound over the values in the grid, replacing \(\delta\) in these bounds with
\(\delta/N\) where \(N\) is the number of grid points.

\paragraph{Existing PAC-Bayes bounds.}
We compare to state-of-the-art PAC-Bayesian bounds (and derived algorithms) for
weighted majority vote classifiers: the First
Order~\citep{langford2003pac}, the Second
Order~\citep{DBLP:conf/nips/MasegosaLIS20},
Binomial~\citep{DBLP:conf/pkdd/LacasseLMT10} (with the number of voters set to \(100\)) and the two
Chebyshev-Cantelli-based~\citep{DBLP:journals/corr/abs-2106-13624} empirical
bounds from categorical-type Gibbs classifiers with parameter
\(\vec{\theta}\), and we refer to these as \emph{FO}, \emph{SO}, \emph{Bin}, \emph{CCPBB} and \emph{CCTND}
respectively (more details are given in \Cref{section:additional-empirical}).
We denote by \emph{f2} the factor two bound derived in~\citet[][Annex A.4]{zantedeschi2021} from Dirichlet majority votes.
All prior distributions for PAC-Bayes bounds, including ours, are set to uniform.
We also refer by the same names to the outputs of optimising these bounds with stochastic gradient descent; details on training and initialisation are given in \Cref{section:additional-empirical}.

\paragraph{Description of figures.}
In \Cref{fig:margins} we compare
\Cref{theorem:main-bound} with the existing margin bound of \Cref{eq:kth-margin} and the
improved \citet{pmlr-v151-biggs22a} bound given in
\Cref{eq:bgplus}. Since \Cref{eq:bgplus} is strictly
better than the original result and the latter was vacuous in almost
all cases considered (see \Cref{section:more-margins-stuff}), we do not include it.  All
datasets are for binary classification as the existing results only cover this
case, and the \(\vec{\theta}\) values considered are the outputs of either the FO- or
f2-optimisation using either the weak or the strong voters described above.
\Cref{fig:pac-bayes} extends this evaluation of \Cref{theorem:main-bound} to improve generalisation results, by applying it to the models optimised with the PAC-Bayes bounds \emph{FO}, \emph{SO}, \emph{Bin} and \emph{f2} as objective. In this case, we consider both binary and multiclass datasets.
In \Cref{fig:optimisation} we directly compare the outputs of optimising state-of-the-art PAC-Bayesian bounds with our optimisation-ready variant result \Cref{theorem:main-bound-stochastic}.
These experiments were carried out on strong voters, as standard in the literature \citep[\emph{e.g.}][]{lorenzen2019pac,DBLP:conf/nips/MasegosaLIS20,DBLP:journals/corr/abs-2106-13624}.

\section{Discussion and conclusion}\label{section:conclusion}

We observe overall that in many cases the existing margin and PAC-Bayes bounds are
insufficient to explain the generalisation observed, while our
new bound is consistently tight, and sometimes sharp (\emph{i.e.} it
approaches the true test error).

\Cref{fig:margins} demonstrates that existing margin bounds can be
insufficient to explain the generalisation observed, which could be
construed as a null result for the ``margins theory''. 
However, our new bound obtains empirically very sharp results in almost all cases, reaffirming to the
theory. 
Note that due to the non-convexity of our bound, the reported values are local minima and can potentially be improved by applying a thorougher search for the optimal $K$, still giving a similarly valid bound.
For instance, simply by enlarging the search space for $K$ our bound drops to $0.36 \pm 0.10$ on \emph{ADULT} with decision stumps as voters, beating existing bounds also in this setting. 
Unlike the existing results, \(\vec{\theta}\) also arises in the complexity
(KL divergence) term and so the bound is not equally tight for every \(\vec{\theta}\)
at fixed margin loss. Further examination of this property could add additional
nuance and perspective to the theory.

When comparing to existing PAC-Bayes bounds in \Cref{fig:pac-bayes}, remarkably
\Cref{theorem:main-bound} is \emph{always} tighter than just using the bound
which is being optimised. We speculate that this arises partially due to the
irreducible factors appearing in those bounds; for example the \emph{FO} or
\emph{f2} bounds can never be tighter than twice the train loss of the associated
Gibbs classifier, while ours has no such limitation. This result is quite
valuable as it demonstrates that \Cref{theorem:main-bound} can be readily used
in an algorithm-free manner: the choice of learning algorithm is up to the practitioner, but the bound
will then often provide an excellent guarantee on the obtained weights
\(\vec{\theta}\).

Finally, in \Cref{fig:optimisation}, our optimisation-friendly variant bound
\Cref{theorem:main-bound-stochastic} is seen to be competitive in terms of test
error while giving an improved-or-equal final bound on all datasets. When
considering the less-common setting of binary stumps (see
\Cref{section:additional-empirical}) we found that sometimes this
objective converged to a sub-optimal local minimum. We speculate that this
arises due to the highly non-convex nature of the objective combined with a
strong \(K\)-inflating gradient signal from the \(O(e^{-K\gamma^{2}})\) term. Thus future
work to improve these results even further
could start with the use of the quasi-convex small-kl relaxation from \citet{DBLP:conf/alt/ThiemannIWS17}. We
note however that this is overall less important than our main results, as both our bounds are still extremely tight
when used in an algorithm-free way and applied to the output of another
algorithm as discussed above.

Overall, we note that in many cases (a majority in \Cref{fig:pac-bayes}) our
main bound of \Cref{theorem:main-bound} is very close to the test set bound and
thus cannot actually be improved any further, with the problem of providing
sharp guarantees based on the training data alone effectively solved in many cases.

\paragraph{Conclusion.}
We obtain empirically very strong generalisation bounds for voting classifiers using margins.
We believe these are highly relevant to the community, since voting-based classifiers and
margin-maximising algorithms are among the most popular and influential in
machine learning.
Dirichlet majority votes have already obtained excellent results in the stochastic setting \citep{zantedeschi2021}, but our new result in \Cref{theorem:dirichlet-margin} showing they are
well-approximated by their mean should open new directions in the more conventional deterministic setting.

Our results also have
practical relevance: for example, in the strong voter machine learning workflow
described above, instead of setting data aside as a test set, this data can be freed up
to learn even stronger voters, since a strong out-of-sample ensemble guarantee can still
be provided even \emph{without} a test set.

In future work we hope to expand these results further to other (non-majority)
voting schemes like those with score-output voters \citep[as in \emph{e.g.}][]{schapireBoostingMarginNew1998}, and ensembles of voters with finite VC
dimension.

\subsection*{Acknowledgements}
The experiments presented in this paper were carried out using the Grid’5000 testbed, supported by a scientific interest group hosted by Inria and including CNRS, RENATER and several Universities as well as other organizations
(see https://www.grid5000.fr). 
F.B. acknowledges the support of the EPSRC grant EP/S021566/1.
V.Z. contributed to this work while being supported from the French National Agency for Research, grant ANR-18-CE23-0015-02.
B.G. acknowledges partial support by the U.S. Army Research Laboratory, U.S. Army Research Office, U.K. Ministry of Defence and the U.K. Engineering and Physical Sciences Research Council (EPSRC) under grant number EP/R013616/1; B.G. also acknowledges partial support from the French National Agency for Research, grants ANR-18-CE40-0016-01 and ANR-18-CE23-0015-02.


{
\small
\bibliography{bibliography}

\begin{thebibliography}{47}
\providecommand{\natexlab}[1]{#1}
\providecommand{\url}[1]{\texttt{#1}}
\expandafter\ifx\csname urlstyle\endcsname\relax
  \providecommand{\doi}[1]{doi: #1}\else
  \providecommand{\doi}{doi: \begingroup \urlstyle{rm}\Url}\fi

\bibitem[Alquier(2021)]{alquier2021userfriendly}
Pierre Alquier.
\newblock User-friendly introduction to {PAC-Bayes} bounds.
\newblock \emph{CoRR}, abs/2110.11216, 2021.
\newblock URL \url{https://arxiv.org/abs/2110.11216}.

\bibitem[Bell and Koren(2007)]{netflix}
Robert~M. Bell and Yehuda Koren.
\newblock Lessons from the netflix prize challenge.
\newblock \emph{SIGKDD Explor. Newsl.}, 9\penalty0 (2):\penalty0 75–79, dec
  2007.
\newblock ISSN 1931-0145.
\newblock \doi{10.1145/1345448.1345465}.
\newblock URL \url{https://doi.org/10.1145/1345448.1345465}.

\bibitem[Biggs and Guedj(2021)]{DBLP:journals/entropy/BiggsG21}
Felix Biggs and Benjamin Guedj.
\newblock Differentiable {PAC-Bayes} objectives with partially aggregated
  neural networks.
\newblock \emph{Entropy}, 23\penalty0 (10):\penalty0 1280, 2021.
\newblock \doi{10.3390/e23101280}.
\newblock URL \url{https://doi.org/10.3390/e23101280}.

\bibitem[Biggs and Guedj(2022{\natexlab{a}})]{biggs2022shallow}
Felix Biggs and Benjamin Guedj.
\newblock Non-vacuous generalisation bounds for shallow neural networks.
\newblock In Kamalika Chaudhuri, Stefanie Jegelka, Le~Song, Csaba
  Szepesv{\'{a}}ri, Gang Niu, and Sivan Sabato, editors, \emph{International
  Conference on Machine Learning, {ICML} 2022, 17-23 July 2022, Baltimore,
  Maryland, {USA}}, volume 162 of \emph{Proceedings of Machine Learning
  Research}, pages 1963--1981. {PMLR}, 2022{\natexlab{a}}.
\newblock URL \url{https://proceedings.mlr.press/v162/biggs22a.html}.

\bibitem[Biggs and Guedj(2022{\natexlab{b}})]{pmlr-v151-biggs22a}
Felix Biggs and Benjamin Guedj.
\newblock On margins and derandomisation in {PAC-Bayes}.
\newblock In Gustau Camps-Valls, Francisco J.~R. Ruiz, and Isabel Valera,
  editors, \emph{Proceedings of The 25th International Conference on Artificial
  Intelligence and Statistics}, volume 151 of \emph{Proceedings of Machine
  Learning Research}, pages 3709--3731. PMLR, 28--30 Mar 2022{\natexlab{b}}.
\newblock URL \url{https://proceedings.mlr.press/v151/biggs22a.html}.

\bibitem[Breiman(1999)]{breiman1999}
Leo Breiman.
\newblock {Prediction Games and Arcing Algorithms}.
\newblock \emph{Neural Computation}, 11\penalty0 (7):\penalty0 1493--1517, 10
  1999.
\newblock ISSN 0899-7667.
\newblock \doi{10.1162/089976699300016106}.
\newblock URL \url{https://doi.org/10.1162/089976699300016106}.

\bibitem[Breiman(2001)]{DBLP:journals/ml/Breiman01}
Leo Breiman.
\newblock Random forests.
\newblock \emph{Mach. Learn.}, 45\penalty0 (1):\penalty0 5--32, 2001.
\newblock \doi{10.1023/A:1010933404324}.
\newblock URL \url{https://doi.org/10.1023/A:1010933404324}.

\bibitem[Catoni(2007)]{catoni2007}
Olivier Catoni.
\newblock \emph{{PAC-Bayesian} Supervised Classification: The Thermodynamics of
  Statistical Learning}.
\newblock Institute of Mathematical Statistics lecture notes-monograph series.
  Institute of Mathematical Statistics, 2007.
\newblock ISBN 9780940600720.
\newblock URL \url{https://books.google.fr/books?id=acnaAAAAMAAJ}.

\bibitem[Chen and Guestrin(2016)]{DBLP:conf/kdd/ChenG16}
Tianqi Chen and Carlos Guestrin.
\newblock Xgboost: {A} scalable tree boosting system.
\newblock In Balaji Krishnapuram, Mohak Shah, Alexander~J. Smola, Charu~C.
  Aggarwal, Dou Shen, and Rajeev Rastogi, editors, \emph{Proceedings of the
  22nd {ACM} {SIGKDD} International Conference on Knowledge Discovery and Data
  Mining, San Francisco, CA, USA, August 13-17, 2016}, pages 785--794. {ACM},
  2016.
\newblock \doi{10.1145/2939672.2939785}.
\newblock URL \url{https://doi.org/10.1145/2939672.2939785}.

\bibitem[Dua and Graff(2017)]{Dua2019}
Dheeru Dua and Casey Graff.
\newblock {UCI} machine learning repository, 2017.
\newblock URL \url{http://archive.ics.uci.edu/ml}.

\bibitem[Dziugaite and Roy(2017)]{dziugaite2017computing}
Gintare~Karolina Dziugaite and Daniel~M Roy.
\newblock Computing nonvacuous generalization bounds for deep (stochastic)
  neural networks with many more parameters than training data.
\newblock \emph{Conference on Uncertainty in Artificial Intelligence 33.},
  2017.

\bibitem[Dziugaite and Roy(2018)]{NIPS2018_8063}
Gintare~Karolina Dziugaite and Daniel~M Roy.
\newblock {Data-dependent {{PAC-Bayes}} priors via differential privacy}.
\newblock In \emph{Advances in Neural Information Processing Systems 31}, pages
  8430--8441. Curran Associates, Inc., 2018.
\newblock URL
  \url{http://papers.nips.cc/paper/8063-data-dependent-pac-bayes-priors-via-differential-privacy.pdf}.

\bibitem[Dziugaite et~al.(2021)Dziugaite, Hsu, Gharbieh, Arpino, and
  Roy]{DBLP:journals/corr/abs-2006-10929}
Gintare~Karolina Dziugaite, Kyle Hsu, Waseem Gharbieh, Gabriel Arpino, and
  Daniel Roy.
\newblock On the role of data in {PAC-Bayes}.
\newblock In Arindam Banerjee and Kenji Fukumizu, editors, \emph{The 24th
  International Conference on Artificial Intelligence and Statistics, {AISTATS}
  2021, April 13-15, 2021, Virtual Event}, volume 130 of \emph{Proceedings of
  Machine Learning Research}, pages 604--612. {PMLR}, 2021.
\newblock URL
  \url{http://proceedings.mlr.press/v130/karolina-dziugaite21a.html}.

\bibitem[Foong et~al.(2021)Foong, Bruinsma, Burt, and
  Turner]{DBLP:conf/nips/FoongBBT21}
Andrew Y.~K. Foong, Wessel~P. Bruinsma, David~R. Burt, and Richard~E. Turner.
\newblock How tight can {PAC-Bayes} be in the small data regime?
\newblock In Marc'Aurelio Ranzato, Alina Beygelzimer, Yann~N. Dauphin, Percy
  Liang, and Jennifer~Wortman Vaughan, editors, \emph{Advances in Neural
  Information Processing Systems 34: Annual Conference on Neural Information
  Processing Systems 2021, NeurIPS 2021, December 6-14, 2021, virtual}, pages
  4093--4105, 2021.
\newblock URL
  \url{https://proceedings.neurips.cc/paper/2021/hash/214cfbe603b7f9f9bc005d5f53f7a1d3-Abstract.html}.

\bibitem[Freund and Schapire(1997)]{DBLP:journals/jcss/FreundS97}
Yoav Freund and Robert~E. Schapire.
\newblock A decision-theoretic generalization of on-line learning and an
  application to boosting.
\newblock \emph{J. Comput. Syst. Sci.}, 55\penalty0 (1):\penalty0 119--139,
  1997.
\newblock \doi{10.1006/jcss.1997.1504}.
\newblock URL \url{https://doi.org/10.1006/jcss.1997.1504}.

\bibitem[Gao and Zhou(2013)]{DBLP:journals/ai/GaoZ13a}
Wei Gao and Zhi{-}Hua Zhou.
\newblock On the doubt about margin explanation of boosting.
\newblock \emph{Artif. Intell.}, 203:\penalty0 1--18, 2013.
\newblock \doi{10.1016/j.artint.2013.07.002}.
\newblock URL \url{https://doi.org/10.1016/j.artint.2013.07.002}.

\bibitem[Germain et~al.(2009)Germain, Lacasse, Laviolette, and
  Marchand]{germainPACBayesianLearningLinear2009}
Pascal Germain, Alexandre Lacasse, Fran\c{c}ois Laviolette, and Mario Marchand.
\newblock {{PAC}}-{{Bayesian}} learning of linear classifiers.
\newblock In \emph{Proceedings of the 26th {{Annual International Conference}}
  on {{Machine Learning}} - {{ICML}} '09}, pages 1--8, {Montreal, Quebec,
  Canada}, 2009. {ACM Press}.
\newblock ISBN 978-1-60558-516-1.
\newblock \doi{10.1145/1553374.1553419}.

\bibitem[Germain et~al.(2015)Germain, Lacasse, Laviolette, Marchand, and
  Roy]{germain2015risk}
Pascal Germain, Alexandre Lacasse, Fran{\c{c}}ois Laviolette, Mario Marchand,
  and Jean{-}Francis Roy.
\newblock Risk bounds for the majority vote: from a {PAC-Bayesian} analysis to
  a learning algorithm.
\newblock \emph{J. Mach. Learn. Res.}, 16:\penalty0 787--860, 2015.
\newblock \doi{10.5555/2789272.2831140}.
\newblock URL \url{https://dl.acm.org/doi/10.5555/2789272.2831140}.

\bibitem[Gr{\o}nlund et~al.(2020)Gr{\o}nlund, Kamma, and
  Larsen]{DBLP:conf/nips/GronlundKL20}
Allan Gr{\o}nlund, Lior Kamma, and Kasper~Green Larsen.
\newblock Margins are insufficient for explaining gradient boosting.
\newblock In Hugo Larochelle, Marc'Aurelio Ranzato, Raia Hadsell,
  Maria{-}Florina Balcan, and Hsuan{-}Tien Lin, editors, \emph{Advances in
  Neural Information Processing Systems 33: Annual Conference on Neural
  Information Processing Systems 2020, NeurIPS 2020, December 6-12, 2020,
  virtual}, 2020.
\newblock URL
  \url{https://proceedings.neurips.cc/paper/2020/hash/146f7dd4c91bc9d80cf4458ad6d6cd1b-Abstract.html}.

\bibitem[Guedj(2019)]{guedj2019primer}
Benjamin Guedj.
\newblock A primer on {PAC-Bayesian} learning.
\newblock \emph{CoRR}, abs/1901.05353, 2019.
\newblock URL \url{http://arxiv.org/abs/1901.05353}.

\bibitem[Kingma and Ba(2015)]{kingma2014adam}
Diederik~P. Kingma and Jimmy Ba.
\newblock Adam: {A} method for stochastic optimization.
\newblock In Yoshua Bengio and Yann LeCun, editors, \emph{3rd International
  Conference on Learning Representations, {ICLR} 2015, San Diego, CA, USA, May
  7-9, 2015, Conference Track Proceedings}, 2015.
\newblock URL \url{http://arxiv.org/abs/1412.6980}.

\bibitem[Lacasse et~al.(2006)Lacasse, Laviolette, Marchand, Germain, and
  Usunier]{DBLP:conf/nips/LacasseLMGU06}
Alexandre Lacasse, Fran{\c{c}}ois Laviolette, Mario Marchand, Pascal Germain,
  and Nicolas Usunier.
\newblock {PAC-Bayes} bounds for the risk of the majority vote and the variance
  of the {Gibbs} classifier.
\newblock In Bernhard Sch{\"{o}}lkopf, John~C. Platt, and Thomas Hofmann,
  editors, \emph{Advances in Neural Information Processing Systems 19,
  Proceedings of the Twentieth Annual Conference on Neural Information
  Processing Systems, Vancouver, British Columbia, Canada, December 4-7, 2006},
  pages 769--776. {MIT} Press, 2006.
\newblock URL
  \url{https://proceedings.neurips.cc/paper/2006/hash/779efbd24d5a7e37ce8dc93e7c04d572-Abstract.html}.

\bibitem[Lacasse et~al.(2010)Lacasse, Laviolette, Marchand, and
  Turgeon{-}Boutin]{DBLP:conf/pkdd/LacasseLMT10}
Alexandre Lacasse, Fran{\c{c}}ois Laviolette, Mario Marchand, and Francis
  Turgeon{-}Boutin.
\newblock Learning with randomized majority votes.
\newblock In Jos{\'{e}}~L. Balc{\'{a}}zar, Francesco Bonchi, Aristides Gionis,
  and Mich{\`{e}}le Sebag, editors, \emph{Machine Learning and Knowledge
  Discovery in Databases, European Conference, {ECML} {PKDD} 2010, Barcelona,
  Spain, September 20-24, 2010, Proceedings, Part {II}}, volume 6322 of
  \emph{Lecture Notes in Computer Science}, pages 162--177. Springer, 2010.
\newblock \doi{10.1007/978-3-642-15883-4\_11}.
\newblock URL \url{https://doi.org/10.1007/978-3-642-15883-4\_11}.

\bibitem[Langford and Seeger(2001)]{Langford01boundsfor}
John Langford and Matthias Seeger.
\newblock Bounds for averaging classifiers, 2001.
\newblock URL
  \url{https://www.cs.cmu.edu/~jcl/papers/averaging/averaging_tech.pdf}.

\bibitem[Langford and Shawe-Taylor(2003)]{langford2003pac}
John Langford and John Shawe-Taylor.
\newblock {{PAC}}-{{Bayes}} \& margins.
\newblock In \emph{Advances in Neural Information Processing Systems}, pages
  439--446, 2003.

\bibitem[Laviolette et~al.(2017)Laviolette, Morvant, Ralaivola, and
  Roy]{laviolette2017risk}
Fran{\c{c}}ois Laviolette, Emilie Morvant, Liva Ralaivola, and Jean{-}Francis
  Roy.
\newblock {Risk upper bounds for general ensemble methods with an application
  to multiclass classification}.
\newblock \emph{Neurocomputing}, 2017.

\bibitem[Letarte et~al.(2019)Letarte, Germain, Guedj, and
  Laviolette]{NIPS2019_8911}
Ga\"el Letarte, Pascal Germain, Benjamin Guedj, and Francois Laviolette.
\newblock Dichotomize and generalize: {{PAC}}-{{Bayesian}} binary activated
  deep neural networks.
\newblock In H.~Wallach, H.~Larochelle, A.~Beygelzimer, F.~dAlch\'e Buc,
  E.~Fox, and R.~Garnett, editors, \emph{Advances in Neural Information
  Processing Systems 32}, pages 6872--6882. {Curran Associates, Inc.}, 2019.

\bibitem[Lorenzen et~al.(2019)Lorenzen, Igel, and Seldin]{lorenzen2019pac}
Stephan~Sloth Lorenzen, Christian Igel, and Yevgeny Seldin.
\newblock On {PAC-Bayesian} bounds for random forests.
\newblock \emph{Mach. Learn.}, 108\penalty0 (8-9):\penalty0 1503--1522, 2019.
\newblock \doi{10.1007/s10994-019-05803-4}.
\newblock URL \url{https://doi.org/10.1007/s10994-019-05803-4}.

\bibitem[Marchal and Arbel(2017)]{marchal2017sub}
Olivier Marchal and Julyan Arbel.
\newblock On the {sub-Gaussianity} of the {Beta} and {Dirichlet} distributions.
\newblock \emph{Electronic Communications in Probability}, 22:\penalty0 1--14,
  2017.

\bibitem[Masegosa et~al.(2020)Masegosa, Lorenzen, Igel, and
  Seldin]{DBLP:conf/nips/MasegosaLIS20}
Andr{\'{e}}s~R. Masegosa, Stephan~Sloth Lorenzen, Christian Igel, and Yevgeny
  Seldin.
\newblock Second order {PAC-Bayesian} bounds for the weighted majority vote.
\newblock In Hugo Larochelle, Marc'Aurelio Ranzato, Raia Hadsell,
  Maria{-}Florina Balcan, and Hsuan{-}Tien Lin, editors, \emph{Advances in
  Neural Information Processing Systems 33: Annual Conference on Neural
  Information Processing Systems 2020, NeurIPS 2020, December 6-12, 2020,
  virtual}, 2020.
\newblock URL
  \url{https://proceedings.neurips.cc/paper/2020/hash/386854131f58a556343e056f03626e00-Abstract.html}.

\bibitem[Maurer(2004)]{DBLP:journals/corr/cs-LG-0411099}
Andreas Maurer.
\newblock A note on the {PAC-Bayesian} theorem.
\newblock \emph{CoRR}, cs.LG/0411099, 2004.
\newblock URL \url{https://arxiv.org/abs/cs.LG/0411099}.

\bibitem[Nielsen(2016)]{nielsen2016tree}
Didrik Nielsen.
\newblock Tree boosting with {XGBoost}: why does {XGBoost} win "every" machine
  learning competition?
\newblock Master's thesis, NTNU, 2016.

\bibitem[Novikoff(1962)]{novikoff62convergence}
A.~B. Novikoff.
\newblock On convergence proofs on perceptrons.
\newblock In \emph{Proceedings of the Symposium on the Mathematical Theory of
  Automata}, volume~12, pages 615--622, New York, NY, USA, 1962. Polytechnic
  Institute of Brooklyn.

\bibitem[Perez-Ortiz et~al.(2021)Perez-Ortiz, Rivasplata, Shawe-Taylor, and
  Szepesvari]{JMLR:v22:20-879}
Maria Perez-Ortiz, Omar Rivasplata, John Shawe-Taylor, and Csaba Szepesvari.
\newblock Tighter risk certificates for neural networks.
\newblock \emph{Journal of Machine Learning Research}, 22\penalty0
  (227):\penalty0 1--40, 2021.
\newblock URL \url{http://jmlr.org/papers/v22/20-879.html}.

\bibitem[Roy et~al.(2011)Roy, Laviolette, and Marchand]{roy2011pac}
Jean{-}Francis Roy, Fran{\c{c}}ois Laviolette, and Mario Marchand.
\newblock From {PAC-Bayes} bounds to quadratic programs for majority votes.
\newblock In Lise Getoor and Tobias Scheffer, editors, \emph{Proceedings of the
  28th International Conference on Machine Learning, {ICML} 2011, Bellevue,
  Washington, USA, June 28 - July 2, 2011}, pages 649--656. Omnipress, 2011.
\newblock URL \url{https://icml.cc/2011/papers/379\_icmlpaper.pdf}.

\bibitem[Schapire et~al.(1998)Schapire, Freund, Bartlett, and
  Lee]{schapireBoostingMarginNew1998}
Robert~E. Schapire, Yoav Freund, Peter Bartlett, and Wee~Sun Lee.
\newblock Boosting the margin: A new explanation for the effectiveness of
  voting methods.
\newblock \emph{The Annals of Statistics}, 26\penalty0 (5):\penalty0
  1651--1686, October 1998.
\newblock \doi{10.1214/aos/1024691352}.

\bibitem[Seeger et~al.(2001)Seeger, Langford, and Megiddo]{seeger2001improved}
Matthias Seeger, John Langford, and Nimrod Megiddo.
\newblock An improved predictive accuracy bound for averaging classifiers.
\newblock In \emph{Proceedings of the 18th International Conference on Machine
  Learning}, number CONF, pages 290--297, 2001.

\bibitem[Shawe{-}Taylor and Hardoon(2009)]{DBLP:journals/jmlr/Shawe-TaylorH09}
John Shawe{-}Taylor and David~R. Hardoon.
\newblock {PAC-Bayes} analysis of maximum entropy classification.
\newblock In \emph{{AISTATS}}, 2009.

\bibitem[Shwartz{-}Ziv and Armon(2022)]{DBLP:journals/inffus/Shwartz-ZivA22}
Ravid Shwartz{-}Ziv and Amitai Armon.
\newblock Tabular data: Deep learning is not all you need.
\newblock \emph{Inf. Fusion}, 81:\penalty0 84--90, 2022.
\newblock \doi{10.1016/j.inffus.2021.11.011}.
\newblock URL \url{https://doi.org/10.1016/j.inffus.2021.11.011}.

\bibitem[Thiemann et~al.(2017)Thiemann, Igel, Wintenberger, and
  Seldin]{DBLP:conf/alt/ThiemannIWS17}
Niklas Thiemann, Christian Igel, Olivier Wintenberger, and Yevgeny Seldin.
\newblock A strongly quasiconvex {PAC-Bayesian} bound.
\newblock In Steve Hanneke and Lev Reyzin, editors, \emph{International
  Conference on Algorithmic Learning Theory, {ALT} 2017, 15-17 October 2017,
  Kyoto University, Kyoto, Japan}, volume~76 of \emph{Proceedings of Machine
  Learning Research}, pages 466--492. {PMLR}, 2017.
\newblock URL \url{http://proceedings.mlr.press/v76/thiemann17a.html}.

\bibitem[Uriot et~al.(2021)Uriot, Izzo, Sim{\~o}es, Abay, Einecke, Rebhan,
  Martinez-Heras, Letizia, Siminski, and Merz]{uriot2021spacecraft}
Thomas Uriot, Dario Izzo, Lu{\'\i}s~F Sim{\~o}es, Rasit Abay, Nils Einecke,
  Sven Rebhan, Jose Martinez-Heras, Francesca Letizia, Jan Siminski, and Klaus
  Merz.
\newblock Spacecraft collision avoidance challenge: design and results of a
  machine learning competition.
\newblock \emph{Astrodynamics}, pages 1--20, 2021.

\bibitem[Viallard et~al.(2021)Viallard, Germain, Habrard, and
  Morvant]{viallard_Cbound}
Paul Viallard, Pascal Germain, Amaury Habrard, and Emilie Morvant.
\newblock Self-bounding majority vote learning algorithms by the direct
  minimization of a tight {PAC-Bayesian C}-bound.
\newblock In \emph{{ECML-PKDD} 2021}, pages 167--183, 2021.

\bibitem[Wang et~al.(2008)Wang, Sugiyama, Yang, Zhou, and
  Feng]{DBLP:conf/colt/Wang08}
Liwei Wang, Masashi Sugiyama, Cheng Yang, Zhi{-}Hua Zhou, and Jufu Feng.
\newblock On the margin explanation of boosting algorithms.
\newblock In Rocco~A. Servedio and Tong Zhang, editors, \emph{21st Annual
  Conference on Learning Theory - {COLT} 2008, Helsinki, Finland, July 9-12,
  2008}, pages 479--490. Omnipress, 2008.
\newblock URL \url{http://colt2008.cs.helsinki.fi/papers/08-Wang.pdf}.

\bibitem[Wu et~al.(2021)Wu, Masegosa, Lorenzen, Igel, and
  Seldin]{DBLP:journals/corr/abs-2106-13624}
Yi{-}Shan Wu, Andr{\'{e}}s~R. Masegosa, Stephan~Sloth Lorenzen, Christian Igel,
  and Yevgeny Seldin.
\newblock {Chebyshev-Cantelli PAC-Bayes-Bennett} inequality for the weighted
  majority vote.
\newblock In Marc'Aurelio Ranzato, Alina Beygelzimer, Yann~N. Dauphin, Percy
  Liang, and Jennifer~Wortman Vaughan, editors, \emph{Advances in Neural
  Information Processing Systems 34: Annual Conference on Neural Information
  Processing Systems 2021, NeurIPS 2021, December 6-14, 2021, virtual}, pages
  12625--12636, 2021.
\newblock URL
  \url{https://proceedings.neurips.cc/paper/2021/hash/69386f6bb1dfed68692a24c8686939b9-Abstract.html}.

\bibitem[Xiao et~al.(2017)Xiao, Rasul, and Vollgraf]{xiao2017}
Han Xiao, Kashif Rasul, and Roland Vollgraf.
\newblock {Fashion-{MNIST}: a Novel Image Dataset for Benchmarking Machine
  Learning Algorithms}.
\newblock \emph{CoRR}, cs.LG/1708.07747, 2017.

\bibitem[Zantedeschi et~al.(2021)Zantedeschi, Viallard, Morvant, Emonet,
  Habrard, Germain, and Guedj]{zantedeschi2021}
Valentina Zantedeschi, Paul Viallard, Emilie Morvant, R{\'{e}}mi Emonet, Amaury
  Habrard, Pascal Germain, and Benjamin Guedj.
\newblock Learning stochastic majority votes by minimizing a {PAC-Bayes}
  generalization bound.
\newblock In Marc'Aurelio Ranzato, Alina Beygelzimer, Yann~N. Dauphin, Percy
  Liang, and Jennifer~Wortman Vaughan, editors, \emph{Advances in Neural
  Information Processing Systems 34: Annual Conference on Neural Information
  Processing Systems 2021, NeurIPS 2021, December 6-14, 2021, virtual}, pages
  455--467, 2021.
\newblock URL
  \url{https://proceedings.neurips.cc/paper/2021/hash/0415740eaa4d9decbc8da001d3fd805f-Abstract.html}.

\bibitem[Zhou et~al.(2019)Zhou, Veitch, Austern, Adams, and
  Orbanz]{DBLP:conf/iclr/ZhouVAAO19}
Wenda Zhou, Victor Veitch, Morgane Austern, Ryan~P. Adams, and Peter Orbanz.
\newblock Non-vacuous generalization bounds at the imagenet scale: a
  {PAC-Bayesian} compression approach.
\newblock In \emph{7th International Conference on Learning Representations,
  {ICLR} 2019, New Orleans, LA, USA, May 6-9, 2019}. OpenReview.net, 2019.
\newblock URL \url{https://openreview.net/forum?id=BJgqqsAct7}.

\end{thebibliography}
}




\clearpage

\appendix


\section{Properties of the Dirichlet distribution}\label{section:about-dirichlet}

The Dirichlet measure has probability density function w.r.t. Lebesgue measure given by:

\[ f(x_1, \dots, x_d; \vec{\alpha}) = \frac{1}{B(\vec{\alpha})} \prod_{i=1}^d x_i^{\alpha_i - 1}\]
where \(B(\vec{\alpha})\) is the multivariate Beta function,
\[ B(\vec{\alpha}) := \frac{\prod_{i=1}^d \Gamma(\alpha_i)}{\Gamma(\sum_{i=1}^d \alpha_d)}.\]

The mean of a Dirichlet is \(\E_{\xi \sim \dirichlet(\vec{\alpha})} \vec{\xi} = \vec{\alpha} / \sum_{i=1}^d \alpha_i\).

The KL divergence between two Dirichlet distributions is the following,
given in \emph{e.g.} \citet{zantedeschi2021}:
\begin{equation*}
  \mathbb{D}_{\dirichlet}(\vec{\alpha}, \vec{\beta}) = \log \frac{B(\vec{\beta})}{B(\vec{\alpha})} + \sum_{i=1}^d (\alpha_i - \beta_i)(\psi(\alpha_i) - \psi(\alpha_0))
  = \log B(\vec{\beta}) - \mathbb{H}_{\dirichlet}(\vec{\alpha}).
\end{equation*}

\section{Additional details on margin bounds}\label{section:more-margins-stuff}

Here we first note the original result from \citet{pmlr-v151-biggs22a} that is
adapted in \Cref{eq:bgplus}; since this is obtained by applying an upper bound
to the inverse small-kl and an additional step, it is strictly looser than the
result we give in \Cref{eq:bgplus}. \citet{pmlr-v151-biggs22a} also uses a
dimension doubling trick to allow negative weights (as they consider only the
binary case), which we remove here to replace the factor \(\log(2d)\) by
\(\log{d}\).

\begin{theorem}\label{theorem:bg-original}
For any margin \(\gamma > 0\), \(\delta \in (0, 1)\), sample size \(m \in \mathbb{N}\), each
of the following results holds with probability at least \(1 - \delta\) over the
sample \(S \sim \mathcal{D}^m\) simultaneously for any \(\vec{\theta} \in \Delta^d\),
\begin{equation}
L_0(\vec{\theta}) \le \hat{L}_{\gamma}(\vec{\theta}) + \sqrt{\frac{C}{m} \cdot \hat{L}_{\gamma}(\vec{\theta})} + \frac{C + \sqrt{C} + 2}{m},
\end{equation}
where \(C := 2\log(2/\delta) + \frac{19}{4}\gamma^{-2} \log d \log m\).
\end{theorem}

\subsection{Definition of the margin}

We here note that the definition of the margin given in
\citet{DBLP:journals/ai/GaoZ13a} and \citet{pmlr-v151-biggs22a} is slightly
different from our own, leading to a scaling of the margin definition by a
factor of one-half. We show this below.

Both the above papers consider prediction functions like
\(F_{\vec{\theta}}(x) = \sum_{i=1}^d \theta_i h_i(x)\) with output set
\(\mathcal{Y} = \{+1, -1\}\). The functions \(h_i(x)\) can be positive or
negative. The margin is defined as as \(y F_{\vec{theta}}(x)\). We translate
this into our equivalent but scaled version as follows:
\begin{equation}\label{eq:margins-equivalence}
y F_{\vec{\theta}}(x) = y\left(\sum_{i : h_i(x) = 1} \theta_i - \sum_{i : h_i(x) = -1} \theta_i\right) = \sum_{i : h_i(x) = y} \theta_i - \sum_{i : h_i(x) = -y} \theta_i
\end{equation}
which is double the margin as we define it. Thus
\(\ell_{\gamma}(\vec{\theta}, x, y) = \1_{M(\vec{\theta}, x, y) \le \gamma} = \1_{y F_{\vec{\theta}} \ge 2\gamma}\)
and the condition on the margin \(yF(x) \ge \sqrt{8/d}\) given in
\citet{DBLP:journals/ai/GaoZ13a} translates to
\(M(\vec{\theta}, x, y) > \sqrt{2/d}\) as we give.

\subsection{Proof of \Cref{theorem:bg-original} and \Cref{eq:bgplus}}

For completeness we provide here short proofs of \Cref{eq:bgplus} and
\Cref{theorem:bg-original}.
The central proposition used in \citet{pmlr-v151-biggs22a} to prove their margin
bound and these results for voting algorithms is the following, proved implicitly there and here
adapted to our setting.

\begin{theorem}[\citet{pmlr-v151-biggs22a}]\label{theorem:categorical-margin}
  Let \(\vec{\theta} \in \Delta^d\) and define \(\rho = \operatorname{Categ}(\vec{\theta})\) and \(\vec{i} \sim \rho^T\) as \(T\) i.i.d. samples from \(\rho\) indexed by \(j \in [T]\). Then for any \(\gamma > 0, T \in \mathbb{N}_{+}\) and \((x, y \in \{+1, -1\})\),
  \begin{align*}
  \ell_0(\vec{\theta}, x, y) \le \E_{\vec{i} \sim \rho^T}\ell^{C}_{\gamma}(\vec{i}, x, y) &+ e^{\frac12 T \gamma^2} \\
  \E_{\vec{i} \sim \rho^T}\ell^{C}_{\gamma}(\vec{i}, x, y)  \le \ell_{\gamma}(\vec{\theta}, x, y) &+ e^{\frac12 T \gamma^2}
  \end{align*}
  where we have defined the margin loss for a sum of Categoricals as
  \(\ell^{\operatorname{C}}_{\gamma}(\vec{i}, x, y) = \1_{y T^{-1} \sum_{t=1}^T h_{i_t}(x) \le \gamma}\).
\end{theorem}

\begin{proof}[Proof of \Cref{eq:bgplus}]
  We apply the PAC-Bayes bound \Cref{theorem:seeger} to
  \(\ell^{C}_{\gamma}\) with \(\rho^T\) as defined in
  \Cref{theorem:categorical-margin} and a uniform prior of the same form,
  \(\pi^T\). We then substitute the results from
  \Cref{theorem:categorical-margin} to show that
  \[
      L_0(\vec{\theta}) \le \smallkl^{-1}\left(\hat{L}_{\gamma}(\vec{\theta}) + e^{-\frac12 T \gamma^2}, \; \frac{\KL(\rho^T, \pi^T) + \log\frac{2 \sqrt{m}}{\delta}}{m} \right)+ e^{-\frac12 T \gamma^2}.
  \]
  With a uniform prior, \(\KL(\rho^T, \pi^T) = T\mathbb{D}_{\operatorname{Cat}}(\theta, d^{-1}\bm{1}) \le T \log d\).
  Substitution of this upper bound and \(T = \ceil*{2 \gamma^{-2} \log m}\) gives the result.
\end{proof}

\begin{proof}[Proof of \Cref{theorem:bg-original}]
  Beginning with \Cref{eq:bgplus}, we relax the ceiling using \(\gamma \le \frac12\)
  and \(m \ge 2\) for a non-vacuous bound to obtain
  \[
  L_0(\vec{\theta}) \le \smallkl^{-1}\left(\hat{L}_{\gamma}(\vec{\theta}) + \frac{1}{m}, \; \frac{C}{2m}\right) + \frac{1}{m}
  \]
  with \(C := 2\log(2/\delta) + \frac{19}{4}\gamma^{-2} \log d \log m\). Then using the small-kl upper bound
  \(\smallkl^{-1}(u, c) \le u + \sqrt{2c u} + 2c\) we have
  \begin{align*}
    L_0(\vec{\theta}) &\le \hat{L}_{\gamma}(\vec{\theta}) + \frac{2}{m} + \sqrt{\left(\hat{L}_{\gamma}(\vec{\theta}) + \frac{1}{m}\right)\frac{C}{m}} + \frac{C}{m} \\
    &\le \hat{L}_{\gamma}(\vec{\theta}) + \frac{2}{m} + \sqrt{\frac{C}{m} \cdot \hat{L}_{\gamma}(\vec{\theta})} + \frac{C + \sqrt{C}}{m} \\
  \end{align*}
  which is the result given.
\end{proof}

\begin{proof}[Proof of \Cref{theorem:categorical-margin}]
  Using the same method as the beginning of the proof of \Cref{eq:margins-equivalence},
  \begin{align*}
    \ell_0(&\vec{\theta}, x, y)  - \E_{\vec{i} \sim \rho^T}\ell^{C}_{\gamma}(\vec{i}, x, y) \\
    &= \E_{\vec{i} \sim \rho^T}[\1_{y F(x) \le 0} - \1_{y T^{-1} \sum_{t=1}^T h_{i_t}(x) \le \gamma}] \\
    &\le \Pr_{\vec{i} \sim \rho^T}\left(\frac12 y (F(x) - T^{-1} \sum_{t=1}^T h_{i_t}(x)) > \frac12\gamma \right) \\
    &\le \exp\left(-\frac12 T \gamma^2\right).
  \end{align*}
  In the last line we used Hoeffding's inequality for a sum of \(T\) random
  variables bounded in \([-1, 1]\). The other side follows using an identical
  method with the margin losses reversed.
\end{proof}

\subsection{Further improvement to the bound}

A question which naturally arises from looking at the proof of \Cref{eq:bgplus}
and \Cref{theorem:bg-original} is whether we can do better by choosing \(T\) in
a more optimal way, rather than just setting it to
\(\ceil*{2\gamma^{-2}\log m}\). We thus prove a bound here which is valid for
the optimal choice of \(T\); in practice this is seen to be slightly tighter
than \Cref{eq:bgplus}, although the improvement from \Cref{theorem:bg-original}
to that result is far greater.

For any \(\vec{\theta} \in \Delta^d\) with probability at least \(1 - \delta\)
over the sample,
\begin{equation}\label{eq:bgplusplus}.
    L_0(\vec{\theta}) \le \inf_{T \in \mathbb{N}_{+}} \left[\smallkl^{-1}\left(\hat{L}_{\gamma}(\vec{\theta}) + e^{-\frac12 T \gamma^2}, \; \frac{T (\log{d} - \mathbb{H}_{\operatorname{Categ}}[\vec{\theta}]) + \log\frac{m}{\delta}}{m} \right)+ e^{-\frac12 T \gamma^2} \right]
\end{equation}

A slightly weaker version of this result, with an extra \(m^{-1}\log(2\sqrt{m})\) term, can be proved from
\[
  L_0(\vec{\theta}) \le \smallkl^{-1}\left(\hat{L}_{\gamma}(\vec{\theta}) + e^{-\frac12 T \gamma^2}, \; \frac{T \log d + \log\frac{2 \sqrt{m}}{\delta}}{m} \right)+ e^{-\frac12 T \gamma^2},
\]
which is shown in the proof of \Cref{eq:bgplus}. We note that the optimal \(T\)
depends on the data only through
\(\hat{L}_{\gamma}(\vec{\theta}) \in \{0, m^{-1}, 2m^{-1}, \dots, 1\}\). The
last possibility gives a trivial bound. A union bound over the \(m\) non-vacuous
possibilities gives \Cref{eq:bgplusplus} with the extra logarithmic factor.

In order to remove this term, we use a slightly more sophisticated argument
applied to a different PAC-Bayes bound (\Cref{theorem:catoni}) given below. This
result uses the function (defined for \(C > 0, p \in [0, 1]\))
\[ \Phi_{C}(p) = -\frac{1}{C}\log(1 - p + pe^{-C})\]
which relates to the small KL (\Cref{theorem:germain-prop}).

\begin{theorem}[\citet{catoni2007}]\label{theorem:catoni}
  Given data distribution \(\mathcal{D}\) on \(\mathcal{X} \times \mathcal{Y}\),
  prior \(P \in \mathcal{M}^{1}(\mathcal{H})\), \(C > 0\) and
  \(\delta \in (0, 1)\), the following hold each with probability
  \(\ge 1 - \delta\) over \(S \sim D^{m}\), for all
  \(Q \in \mathcal{M}^{1}(\mathcal{H})\)
  \begin{equation*}
    \E_{h \sim Q}L(h) \le \Phi_{C}^{-1} \left( \E_{h \sim Q}\hat{L}(h) + \frac{\KL(Q, P) + \log\frac{1}{\delta}}{Cm} \right)
  \end{equation*}
\end{theorem}

\begin{theorem}[\citet{germainPACBayesianLearningLinear2009}, Proposition 2.1]\label{theorem:germain-prop}
  For any \(0 \le q \le p < 1\),
  \[ \sup_{C > 0} \left[ C\Phi_{C}(p) - Cq \right] = \smallkl(q, p). \]
\end{theorem}

\begin{proof}[Proof of \Cref{eq:bgplusplus}]
  We substitute \Cref{theorem:categorical-margin} into \Cref{theorem:catoni} with the categorical loss and
  a uniform prior, \(\pi^T\). and KL upper bound as in the above proof. as we obtain for any data-independent
  \(C > 0, T \in \mathbb{N}_{+}, \gamma > 0\) that
  \[
    L_0(\vec{\theta}) - e^{-\frac12 T \gamma^2} \le \Phi_{C}^{-1} \left( \frac{k}{m} +  e^{-\frac12 T \gamma^2} + \frac{T\log d + \log\frac{1}{\delta}}{Cm} \right).
  \]
  where \(k := m\hat{L}_{\gamma}(\vec{\theta})\) is the number of margin errors.

  Since the only quantity on the left hand side in this bound unknown before we
  see data is the value of \(k\), there exists a \(C_k\) dependent on the value
  of \(k\) that optimises the bound, and a \(T_k\) that depends on this pair.
  Since there are only \(m\) such values giving non-vacuous bounds (\(k = m\) is
  trivially vacuous), we can apply a union bound over all these bounds with
  \(\delta = \delta/m\) to give the following with probability
  \(\ge 1 - \delta\):
  \[
    L_0(\vec{\theta}) \le \min_{T \in \mathbb{N}_{+}} \min_{C > 0}\left[ e^{-\frac12 T \gamma^2} + \Phi_{C}^{-1} \left( \frac{k}{m} +  e^{-\frac12 T \gamma^2} + \frac{T\log d + \log\frac{m}{\delta}}{Cm} \right) \right].
  \]
  Applying the inversion of \Cref{theorem:germain-prop} gives the second result.
\end{proof}

\subsection{Comparison of margin bounds}

\begin{figure}[H]
\centering
\includegraphics[width=\textwidth]{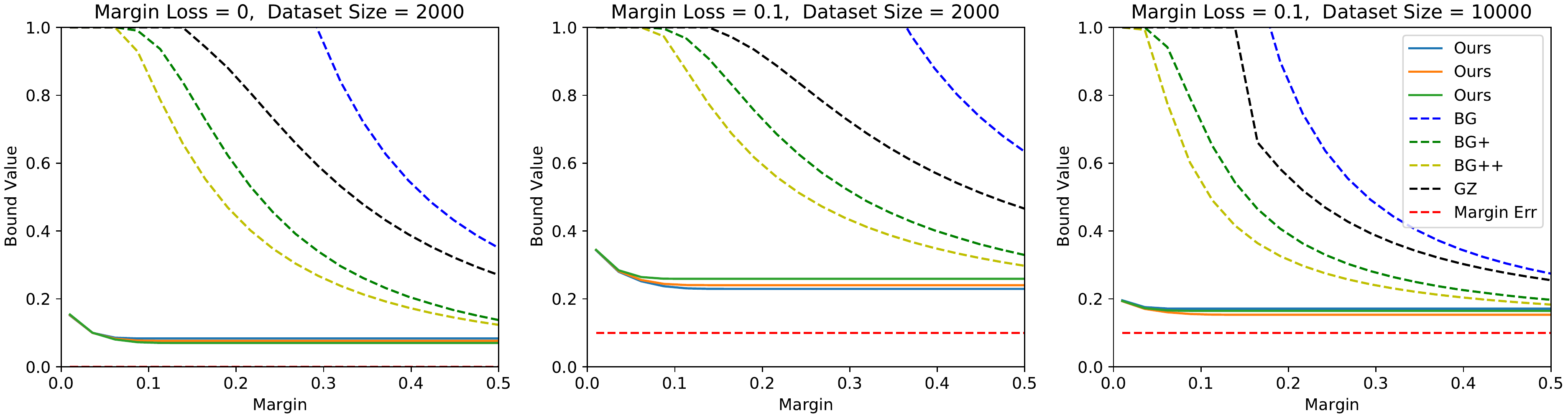}
\caption{Values of different bounds versus margin at margin error \(\hat{L}_{\gamma}\) (\(0\) or \(0.1\)). Dimension \(d = 100\), probability \(\delta = 0.5\) and dataset size \(m\) (\(2000\) or \(10000\)) are also fixed. The bounds are  \Cref{theorem:main-bound} (\textbf{ours}) with three different samples \(\vec{\theta} \sim \operatorname{Uniform}(\Delta^d)\), compared with the margin bounds of \Cref{theorem:bg-original} (\textbf{BG}), \Cref{eq:bgplus} (\textbf{BG+}), \Cref{eq:bgplusplus} (\textbf{BG++}), \Cref{eq:kth-margin} (\textbf{GZ}), and the margin error \(\hat{L}_{\gamma}\).}
\label{fig:margins-comparison}
\end{figure}

In \Cref{fig:margins-comparison} we compare the various bounds given above in a
non-experimental way, fixing the margin loss \(\hat{L}_{\gamma}\) to a
particular value and seeing how the bounds change if that value of the loss is
achieved for different values of the margin \(\gamma \in (0, 0.5)\). Since
(uniquely among the bounds), the value of \(\vec{\theta}\) appears in our bound
\Cref{theorem:main-bound}, we show three different sampled possible values,
drawn uniformly from the simplex.

The results for ``categorical''-based bounds demonstrate that the refined bounds
\Cref{eq:bgplus,eq:bgplusplus} are much tighter that the result as given in
\Cref{theorem:bg-original} by \citet{pmlr-v151-biggs22a}. Both these refinements
are also tighter than \Cref{eq:kth-margin} from
\citet{DBLP:journals/ai/GaoZ13a}. We used \Cref{eq:bgplus} in the main paper
because it is closer to an exiting result (as it appears in the proof from
\citealp{pmlr-v151-biggs22a}), and is not much worse than the refinement
\Cref{eq:bgplusplus}, particularly when compared to our far stronger new result
\Cref{theorem:main-bound}.

This figure also shows that, at least for some values of \(\vec{\theta}\), this
new bound can be far tighter than all the existing bounds. One interesting facet
of this is that the bound is improved very little for \(\gamma\) above a certain
point, quite a different behaviour to the other bounds. Empirically this was
seen too in our other experiments, with the optimised \(\gamma\) often being
quite small. Of course, for some values of \(\vec{\theta}\) this bound will be
weaker, but we observe the same kind of results in our main experimental
results, where this is a learned value.

\section{Additional experimental details and evaluations}\label{section:additional-empirical}

\paragraph{Dataset descriptions.}\label{app:datasets}

We provide the description of the classification datasets considered in our empirical evaluation.
\begin{description}
    \item[\hspace{0.5cm}\emph{Haberman} (UCI)] prediction of survival of $n=306$ patients who had undergone surgery from $d=3$ anonymized features.
    \item[\hspace{0.5cm}\emph{TicTacToe} (UCI)] determination of a win for player $x$ at TicTacToe game of any of the $n=958$ board configurations ($d=9$ categorical states).
  \item[\hspace{0.5cm}\emph{Mushrooms} (UCI)] prediction of edibility of $n=8,124$ mushroom sample, given their $d=22$ categorical features describing their aspect.
  \item[\hspace{0.5cm}\emph{Adult} (LIBSVM a1a)] determining whether a person earns more than 50K a year ($n=32,561$ people and $d=123$ binary features).
  \item[\hspace{0.5cm}\emph{CodRNA} (LIBSVM)] detection of non-coding RNAs among $n=59,535$ instances and from $d=$ features.
  \item[\hspace{0.5cm}\emph{Pendigits} (UCI)] recognition of hand-written digits ($10$ classes, $d=9$ features and $n= 12,992$).
  \item[\hspace{0.5cm}\emph{Protein} (LIBSVM)] $d=357$ features, $n=24,387$ instances and $3$ classes.
  \item[\hspace{0.5cm}\emph{Sensorless} (LIBSVM)] prediction of motor condition ($n=58,509$ instances and $11$ classes), with intact and defective components, from $d=48$ features extracted from electric current drive signals.
  \item[\hspace{0.5cm}\emph{MNIST} (LIBSVM)] prediction of hand-written digits ($n=70,000$ instances and $10$ classes) from $d=28 \times 28$ gray-scale images.
  \item[\hspace{0.5cm}\emph{Fashion-MNIST} (Zalando)] prediction of cloth articles ($n=70,000$ instances and $10$ classes) from $d=28 \times 28$ gray-scale images.
\end{description}
In all experiments, we convert all categorical features to numerical using an ordinal encoder and we standardize all features using the statistics of the training set. 

\paragraph{Baseline descriptions.}
We report the generalization bounds of the literature used for training weighted majority vote classifiers in our comparison.
We additionally note:
$\categorical(\vec{\theta})$ the categorical distribution over the base classifiers (with $\theta_i$ the weight associated to voter \(h_i \in \mathcal{H}\)), and $\mathbb{D}_{\categorical}(\vec{\theta}), \vec{\pi})$ the KL-divergence between two categorical distribution with parameters $\vec{\theta}$ and $\vec{\pi}$;
$\ell_{\text{TND}}(h, h', x, y) := \1_{h(x) \ne y \land h'(x) \ne y}$ the tandem loss proposed in~\citet{DBLP:conf/nips/MasegosaLIS20} and \(\hat{L}_{\text{TND}}(h, h') := \E_{(x, y) \sim \operatorname{Uniform}(S)} \ell_{\text{TND}}(h, h', x, y)\) its in-sample estimate;
$\ell_{\text{Bin}}(\vec{\theta}, N, x, y) := \sum_{k=\frac{N}{2}}^N {N \choose k} M(\vec{\theta}, x, y)^k (1-M(\vec{\theta}, x, y))^{(N-k)}$ the probability that among $N$ voters randomly drawn from $\categorical(\vec{\theta})$ at least $\frac{N}{2}$ of them are incorrect, as defined in \citet{DBLP:conf/pkdd/LacasseLMT10}.

\begin{itemize}
  \item First Order~\citep[FO,][]{langford2003pac}:\\
  For any \(\mathcal{D} \in \measure(\mathcal{X} \times \mathcal{Y})\), \(m \in \mathbb{N}_{+}\), \(\delta \in (0, 1)\), and prior \(\vec{\pi} \in \Delta^d\), with probability at least \(1 - \delta\) over the sample \(S \sim \mathcal{D}^m\) simultaneously for every \(\vec{\theta} \in \Delta^d\),

  \begin{equation*}
    L_0(\vec{\theta}) \le 2 \smallkl^{-1} \left(\E_{h \sim \categorical(\vec{\theta})}\hat{L}_0(h), \; \frac{\mathbb{D}_{\categorical}(\vec{\theta}, \vec{\pi}) + \log\frac{2\sqrt{m}}{\delta}}{m} \right).
  \end{equation*}

  \item Second Order~\citep[SO,][]{DBLP:conf/nips/MasegosaLIS20}:\\
  For any \(\mathcal{D} \in \measure(\mathcal{X} \times \mathcal{Y})\), \(m \in \mathbb{N}_{+}\), \(\delta \in (0, 1)\), and prior \(\vec{\pi} \in \Delta^d\), with probability at least \(1 - \delta\) over the sample \(S \sim \mathcal{D}^m\) simultaneously for every \(\vec{\theta} \in \Delta^d\),

  \begin{equation*}
    L_0(\vec{\theta}) \le 4 \smallkl^{-1} \left(\E_{h \sim \categorical(\vec{\theta}), h' \sim \categorical(\vec{\theta})}\hat{L}_{\text{TND}}(h, h'), \; \frac{2 \mathbb{D}_{\categorical}(\vec{\theta}, \vec{\pi}) + \log\frac{2\sqrt{m}}{\delta}}{m} \right).
  \end{equation*}

  \item Binomial~\citep[Bin,][]{DBLP:conf/pkdd/LacasseLMT10}:\\
  For any \(\mathcal{D} \in \measure(\mathcal{X} \times \mathcal{Y})\), \(m \in \mathbb{N}_{+}\), \(N \in \mathbb{N}_{+}\), \(\delta \in (0, 1)\), and prior \(\vec{\pi} \in \Delta^d\), with probability at least \(1 - \delta\) over the sample \(S \sim \mathcal{D}^m\) simultaneously for every \(\vec{\theta} \in \Delta^d\),

  \begin{equation*}
    L_0(\vec{\theta}) \le 2 \smallkl^{-1} \left(\E_{(x, y) \sim \operatorname{Uniform}(S)}\ell_{\text{Bin}}(\vec{\theta}, N, x, y), \; \frac{N \mathbb{D}_{\categorical}(\vec{\theta}, \vec{\pi}) + \log\frac{2\sqrt{m}}{\delta}}{m} \right).
  \end{equation*}

  \item Chebyshev-Cantelli tandem loss bound~\citep[CCTND,][Theorem 12]{DBLP:journals/corr/abs-2106-13624};


  \item Chebyshev-Cantelli tandem loss bound with an offset~\citep[CCPBB,][Theorem 15]{DBLP:journals/corr/abs-2106-13624};

  \item Dirichlet Factor-Two~\citep[f2,][]{zantedeschi2021}: \\
  For any \(\mathcal{D} \in \measure(\mathcal{X} \times \mathcal{Y})\), \(m \in \mathbb{N}_{+}\), \(\delta \in (0, 1)\), and prior \(\vec{\beta} \in \Re_{+}^d\), with probability at least \(1 - \delta\) over the sample \(S \sim \mathcal{D}^m\) simultaneously for every \(\vec{\theta} \in \Delta^d\) and \(K > 0\),

  \begin{equation*}
    L_0(\vec{\theta}) \le 2 \smallkl^{-1} \left(\E_{\vec{\xi} \sim \dirichlet(K \vec{\theta})}\hat{L}(\vec{\xi}), \; \frac{\mathbb{D}_{\dirichlet}(K \vec{\theta}, \vec{\beta}) + \log\frac{2\sqrt{m}}{\delta}}{m} \right).
  \end{equation*}

\end{itemize}

\paragraph{Optimisation of PAC-Bayesian bounds.}
To optimize the baselines \emph{CCPBB} and \emph{CCTND}, we rely on the code released by its authors~\footnote{\url{https://github.com/StephanLorenzen/MajorityVoteBounds/tree/44cec987865ddce01cd27076019394538cee85ca/NeurIPS2021}}, with the Gradient Descent option and building random forests as described in our main text.
When optimising the PAC-Bayesian bounds \emph{FO}, \emph{SO}, \emph{Bin}, \emph{f2} and ours, we initialize \(\vec{\theta}\)'s to be uniform, i.e. \(\theta_i = 1/d\), and $K = 2$.
We then optimise the posterior parameters of the method ($\alpha = K \theta$ for Dirichlet, and $\theta$ for Categorical distributions) with the Adam optimiser~\citep{kingma2014adam} with running average coefficients $(0.9, 0.999)$, batch size equal to $100$ and learning rate set to $0.1$.
All methods are run for a maximum of $100$ epochs with patience of $25$ epochs for early stopping and a learning rate scheduler reducing it by a factor of $10$ with $2$ epochs patience.

At each run of an algorithm, we randomly split a dataset into training and test sets of sizes $80\%-20\%$ respectively, and optimise/evaluate the bounds only with the half of the training set that was not used for learning the voters (in the case of data-dependent ones).
Note that we do not make use of a validation set, as we use the risk certificates as estimate of the test error for model selection.
Finally, we report the value of Seeger's "small-kl" bound of~\Cref{theorem:seeger}, even when a different type of bound has been optimised (\emph{e.g.} for the \emph{CCPBB} and \emph{CCTND} baselines), and we average all results over $5$ different trials.

\paragraph{Margin bound comparison.}
Given a pre-trained model, hence fixed $\theta$ and initial $K_{\mathit{init}}$ (which is different from $1.$ only for the models trained via Dirichlet bounds), we search for its optimal risk certificate by evaluating a given bound at $1,000$ values of $\gamma$, spaced evenly on a log scale with base $10$ and in the interval $[10^{-4}, 0.5)$.
For our margin bound, for each of these $\gamma$ values we also optimise $K \in [K_{\mathit{init}}, K_{\mathit{init}} \; 2^{16}]$ using the golden-section search technique to obtain the tightest upper bound.
Notice that this does not add significant computational overhead to the search.
Also for these experiments, the bounds are evaluated with the portion of training data that was not used for learning the voters.

\paragraph{Compute.} All experiments were run on a virtual machine with $16$ vCPUs and $128Gb$ of RAM.

\subsection{Additional results}
\begin{figure}
\centering
\includegraphics[width=\textwidth]{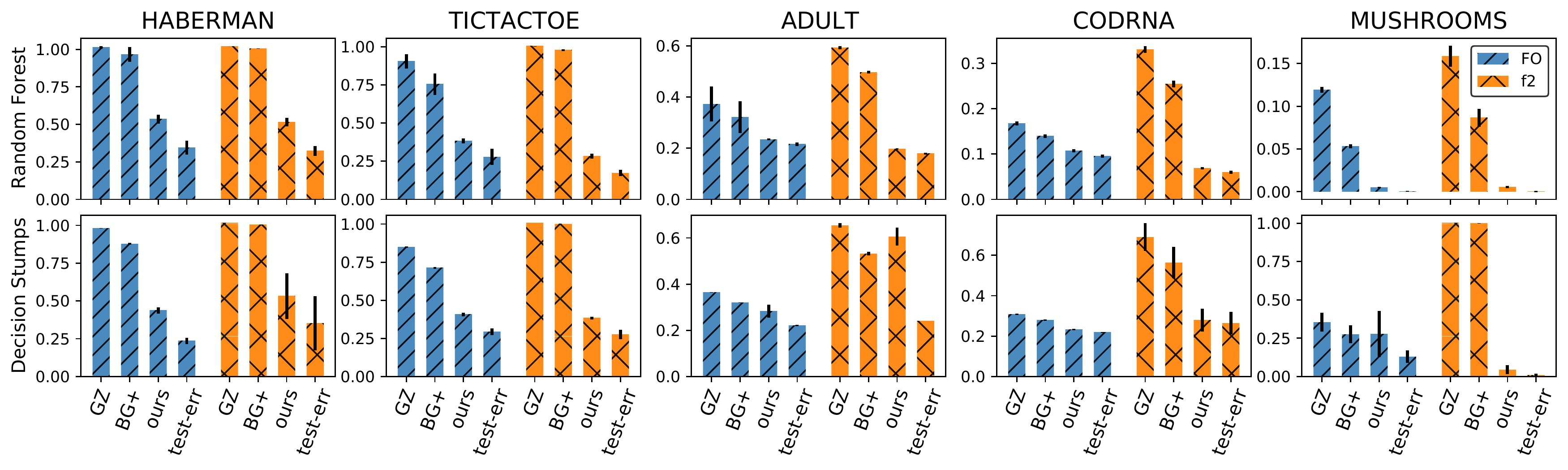}
\caption{\Cref{theorem:main-bound} (\textbf{ours}) compared with the margin bounds of \Cref{eq:bgplus} (\textbf{BG+}), \Cref{eq:kth-margin} (\textbf{GZ}), and the test error. Settings are \emph{rf} (first row) and \emph{stumps} (second row) on the given datasets, with \(\vec{\theta}\) output by optimising either \emph{FO} or \emph{f2} (first and second column groupings respectively).}
\label{fig:margin-rescaled}
\end{figure}

\begin{figure}
\centering
\includegraphics[width=\textwidth]{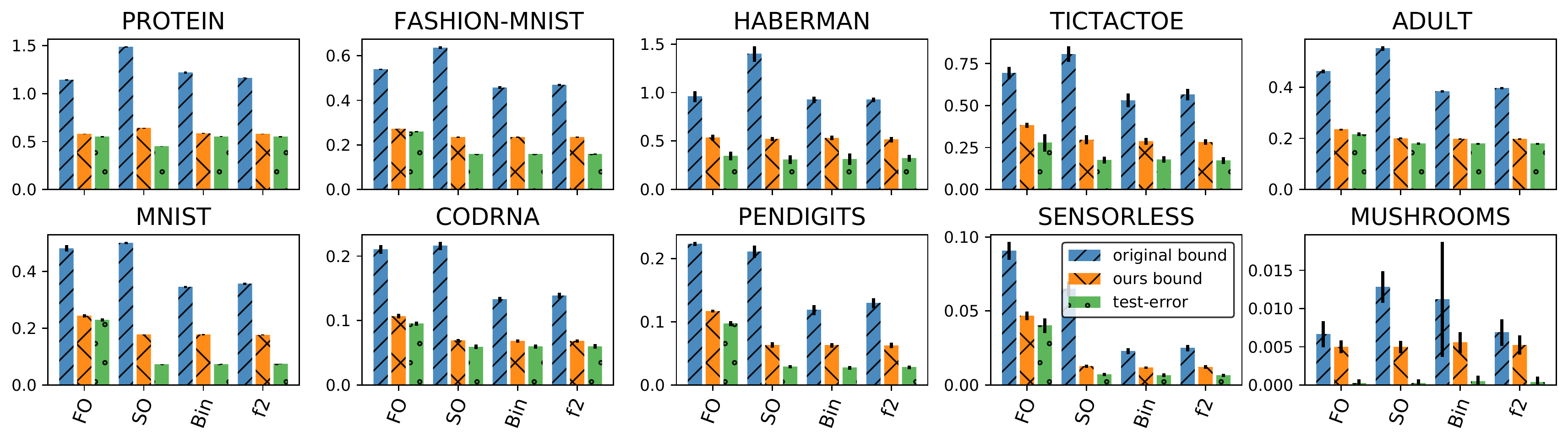}
\caption{\Cref{theorem:main-bound} (\emph{our bound}) compared with the bounds of  \emph{FO}, \emph{SO}, \emph{Bin} or \emph{f2} (\emph{original bound}), and test errors. For each column grouping, \(\vec{\theta}\) is the output from optimising the corresponding PAC-Bayes bound for \emph{rf} on the given dataset.}
\label{fig:original-rescaled}
\end{figure}

\begin{figure}
\centering
\includegraphics[width=\textwidth]{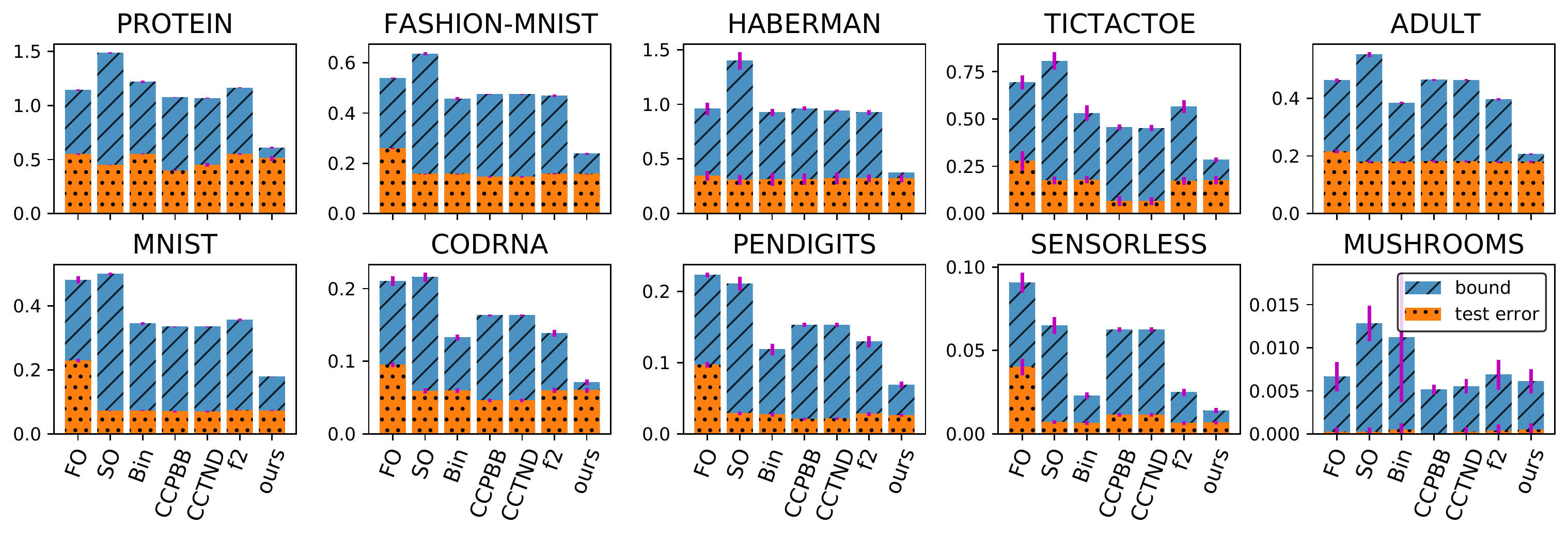}
\caption{\Cref{theorem:main-bound-stochastic} (\textbf{ours}) as optimisation objective compared to other PAC-Bayes results (\emph{FO}, \emph{SO}, \emph{Bin}, \emph{CCPBB}, \emph{CCTND} and \emph{f2}) as objectives, with a Random Forest as set of voters.}
\label{fig:rf-rescaled}
\end{figure}

\begin{figure}
\centering
\includegraphics[width=0.7\textwidth]{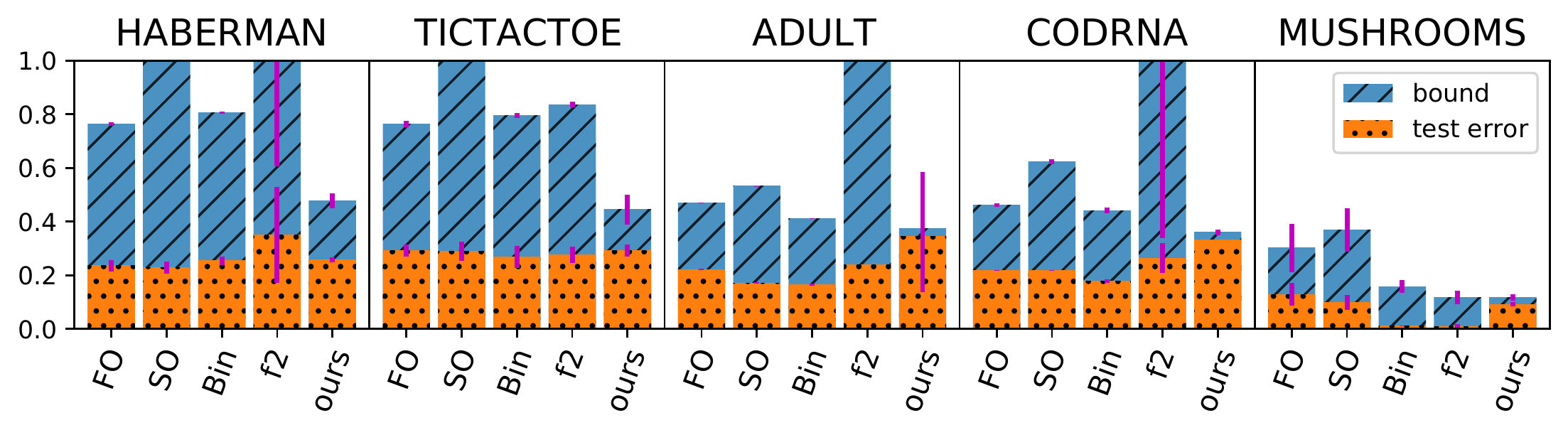}
\caption{\Cref{theorem:main-bound-stochastic} (\textbf{ours}) as optimisation objective compared to other PAC-Bayes results (\emph{FO}, \emph{SO}, \emph{Bin}, \emph{f2}) as objectives, with decision stumps as voters.}
\label{fig:stumps-rescaled}
\end{figure}

In \Cref{fig:margin-rescaled}, \Cref{fig:original-rescaled} and \Cref{fig:rf-rescaled} we report the results from \Cref{fig:margins}, \Cref{fig:pac-bayes} and \Cref{fig:optimisation} in the main text.
Here we deploy a different scale per dataset so that they can be easily read, also when the bounds and test errors are very small.
Additionally, in \Cref{fig:stumps-rescaled} we provide the test errors and risk certificates obtained by optimising the generalization bounds with decision stumps as voters.
Although our certificates are always the tightest, we found that in some cases our method converges to sub-optimal solutions. 
We speculate that this arises due to the highly non-convex nature of the objective combined with a
strong \(K\)-inflating gradient signal from the \(O(e^{-K\gamma^{2}})\) term. Thus future
work to improve these results even further
could start with the use of the quasi-convex small-kl relaxation from \citet{DBLP:conf/alt/ThiemannIWS17}. We
note however that this is overall less important than our main results, as both our bounds are still extremely tight
when used in an algorithm-free way and applied to the output of another
algorithm.

\end{document}